\documentclass{article}

\usepackage[preprint]{neurips_2025}

\usepackage[utf8]{inputenc} 
\usepackage[T1]{fontenc}    
\usepackage{hyperref}       
\usepackage{url}            
\usepackage{booktabs}       
\usepackage{amsfonts}       
\usepackage{nicefrac}       
\usepackage{microtype}      
\usepackage{xcolor}         
\usepackage{amsmath}
\usepackage{amssymb}
\usepackage{amsfonts}
\usepackage{listings}
\usepackage{upquote}  
\usepackage{xcolor,colortbl}
\usepackage{graphicx}
\usepackage{multirow}

\usepackage{algorithm}
\usepackage{upquote}  
\usepackage{xcolor,colortbl}
\lstdefinestyle{overleaf}{
    backgroundcolor=\color[rgb]{0.95,0.95,0.92},   
    commentstyle=\color[rgb]{0,0.6,0},
    keywordstyle=\color{magenta},
    numberstyle=\tiny\color[rgb]{0.5,0.5,0.5},
    stringstyle=\color[rgb]{0.58,0,0.82},
    basicstyle=\ttfamily\footnotesize,
    breakatwhitespace=false,         
    breaklines=true,                 
    captionpos=b,                    
    keepspaces=true,                 
    numbers=left,                    
    numbersep=5pt,                  
    showspaces=false,                
    showstringspaces=false,
    showtabs=false,                  
    tabsize=2
}
\lstdefinestyle{mocov3}{
  backgroundcolor=\color{white},
  basicstyle=\fontsize{7.5pt}{7.5pt}\ttfamily\selectfont,
  columns=fullflexible,
  breaklines=true,
  captionpos=b,
  commentstyle=\fontsize{7.5pt}{7.5pt}\color[rgb]{0.25,0.5,0.5},
  keywordstyle=\fontsize{7.5pt}{7.5pt}\color[rgb]{0.85,0.18,0.50},
}

\newcommand{\img}{I}

\title{Test-Time Training Done Right}

\newcommand*{\affmark}[1][*]{\textsuperscript{#1}}

\author{%
  Tianyuan Zhang\affmark[1]~~~
  Sai Bi\affmark[2]~~~
  Yicong Hong\affmark[2]~~~
  Kai Zhang\affmark[2]~~~
  Fujun Luan\affmark[2]~~~ \\
  \textbf{Songlin Yang\affmark[1]}~~~
  \textbf{Kalyan Sunkavalli\affmark[2]}~~~
  \textbf{William T. Freeman\affmark[1]}~~~
  \textbf{Hao Tan\affmark[2]}~~~ \\ \\
  $^1$Massachusetts Institute of Technology~~~~~~
    $^2$Adobe Research\\
}

\begin{document}

\maketitle

\maketitle



\newcommand{\methodname}{LaCT}

\def\capfont{\normalfont\small}

\def\capfontfootnotesize{\normalfont\footnotesize}
\def\capfontscriptsize{\normalfont\scriptsize}

\begin{abstract}

Test-Time Training (TTT) models context dependencies by adapting part of the model's weights (often referred to as fast weights) at inference time. This adapted fast weight, similar to recurrent states in RNNs, stores temporary memories of past tokens in the current sequence. Existing TTT methods have struggled to demonstrate effectiveness in handling long-sequence data, due to their computational inefficiency on modern GPUs. The TTT layers in
many of these approaches operate with extremely low FLOPs utilization (often below 5\%) because they deliberately apply small online mini-batch sizes (e.g., updating fast weights every 16 or 64 tokens).
Moreover, a small mini-batch implies fine-grained block-wise causal dependencies in the data, making them unsuitable for data beyond 1D ordered sequences, like sets or N-dimensional grids such as images or videos.
In contrast, we pursue the opposite direction by proposing an extremely large chunk update, ranging from 2K to 1M tokens across tasks of varying modalities, which we refer to as Large Chunk Test-Time Training (LaCT). 
This approach improves hardware utilization by orders of magnitude, and more importantly, facilitates scaling of nonlinear state size (up to 40\% of model parameter size), hence
substantially improving state capacity, all without requiring cumbersome and error-prone custom kernel implementations.
It also allows easy integration of sophisticated optimizers like Muon for online memory updates.
We validate our approach across diverse data modalities and tasks, including novel view synthesis from image sets, language models, and auto-regressive video diffusion models. 
Our approach can scale up to 14-billion-parameter auto-regressive video diffusion models handling sequences of up to 56K tokens. 
In our longest sequence experiment, we perform novel view synthesis with more than one million context length. Our results highlight the computational and performance benefits of large-chunk test-time training, paving the way for more efficient and scalable long-context sequence modeling. We hope that this work will inspire and accelerate new research in the field of long-context modeling and test-time training.  See visual results on project website \url{https://tianyuanzhang.com/projects/ttt-done-right/}.

\end{abstract}
\section{Introduction} \label{sec:Introduction}

The demand for handling long contexts is rapidly growing. While softmax attention~\cite{vaswani2017attention} has become the de facto solution for modeling various types of data, its computational cost grows quadratically with sequence length, motivating extensive research into more 
efficient long-context modeling.

Recently, Test-Time Training (TTT)~\cite{sun2024learning} has emerged as a promising approach for efficient sub-quadratic sequence modeling. TTT extends the concept of recurrent states in RNNs to a small, online-adapted sub-network.
The parameters of this sub-network also referred to as fast weight~\cite{schlag2021linear}, as they are rapidly adapted online via self-supervised objectives to memorize in-context information. 
Numerous recent studies~\cite{wang2025testtimeregressionunifyingframework,behrouz2024titans,behrouz2025itsconnectedjourneytesttime, karami2025lattice} have explored various online objectives, optimizers, and architectures for fast weight networks.

Despite these efforts, existing TTT methods struggle to scale effectively to long contexts, primarily due to extremely low hardware utilization in their TTT layers (often below 5\% peak FLOPS on modern GPUs). 
This inefficiency is because of the usage of
small mini-batch sizes, i.e. updating fast weights every token or every 16 to 64 tokens, which is conventionally assumed to be more effective for in-context learning.
Such small mini-batch results in poor parallelism and low compute intensity, and presents significant challenges for hardware-efficient 
implementation, especially when using large, nonlinear fast weights, making it difficult to achieve non-trivial (above 10\%) FLOPs utilization.

In this paper, we adopt the opposite strategy and introduce Large Chunk Test-Time Training (\methodname). 
\methodname{} leverages extremely large chunk (from 2048 to 1M tokens) as the basic unit to update the fast weight.
Since the tokens within each large chunk are treated as an unordered set, we further integrate window attention
into \methodname{} to capture local dependencies within the chunk.
\methodname{} significantly enhances parallelism, leading to substantially improved GPU utilization (up to 70\% on NVIDIA A100s)  with just a few dozen lines of pure PyTorch code (see Appendix~\ref{sec:ttt_pseudocode}).
This efficiency enables the scaling of non-linear fast weights to enhance the memory capacity. And simple implementation allows easy integration of more effective test-time optimizers, such as Muon~\cite{jordan2024muon}.

Furthermore, \methodname's large-chunk design is also natural to model diverse N-dimensional data as we can align chunk-size with the internal structure of the data (e.g., grouping tokens within an image or consecutive video frames as a chunk).

We extensively  validate \methodname{} on three tasks spanning different modalities and data structures: 
\begin{itemize}
\item \em{Novel View Synthesis}.
Our model is capable of processing up to $128$ input images 
at a resolution of $960\!\times\!536$ leading to a maximum of $1$M tokens, and  outperforms 3D Gaussian Splatting~\cite{kerbl20233d} in terms of rendering quality under such input scale.
\item \em{Language Modeling}.
Our model achieves competitive performance compared to SoTA methods such as DeltaNet~\cite{yang2024parallelizing}, even though a chunk structure is not explicitly present in language data.
\item \em{Autoregressive Video Diffusion}. 
We adapt a 14-billion-parameter bidirectional video diffusion transformer into an autoregressive model by incorporating \methodname{} with sliding window attention. This adapted model generates consistent videos up to 56,000 visual tokens.
\end{itemize}

To summarize, our approach establishes an efficient, scalable, and highly performant framework for long 
sequence modeling across diverse modalities. 
By removing the dependency on low-level, hardware-specific implementations, \methodname{} enables broader exploration of the architectural design space. We believe this can democratize research in efficient long-context modeling and inspire the development of more novel and effective designs.

\section{Preliminary} \label{sec:related}

\subsection{Test-Time Training}

Consider a one-dimensional sequence of $N$ tokens
$\mathbf{x} = [x_1, x_2, \dots, x_N]$, where each token $x_i \in \mathbb{R}^d$. 
Following attention formulation, 
each input tokens $x_i$ is projected into query ($q_i$), key ($k_i$), and value ($v_i$) vectors. 
For clarity, we assume all these vectors $q_i, k_i, v_i \in \mathbb{R}^d$.

Test-Time Training (TTT)~\cite{sun2024learning} introduces a neural network with rapidly adaptable weights---called \textit{fast weights}~\cite{schlag2021linear}---that are updated during 
both training and inference to dynamically store context information. 
This contrasts with the \textit{slow weights} (i.e., model parameters) that are frozen during inference.
Formally, TTT defines fast weights in the form of a neural network:
$f_W(\cdot): \mathbb{R}^d \rightarrow \mathbb{R}^d$ parameterized by the fast weights $W$,  and it involves two primary operations:
\begin{equation}
\textbf{Update operation:} \quad W \leftarrow W - \eta \nabla_{W} \mathcal{L}\big(f_W(k), v\big)
\label{eq:ttt_update}
\end{equation}
where $\mathcal{L}(\cdot,\cdot)$ is a loss function between the transformed key  $f_W(k)$ and the value $v$, commonly Mean Squared Error, designed to encourage the network to associate keys with corresponding values. 
$\eta$ is the learning rate.
Intuitively, this learning objective is to encode the KV cache into a neural memory with fixed state size as \emph{accurate} as possible~\cite{wang2025testtimeregressionunifyingframework}.
\begin{equation}
\textbf{Apply operation:}\quad \quad \quad
o = f_W(q),\quad 
\label{eq:ttt_apply}
\end{equation}
where the updated fast weights $W$ are used to compute the output vector $o$ given the query $q$.
The per-token TTT layer iteratively perform the update and apply operations on each token $x_i$ in sequence.

\subsection{Challenges in Efficient Implementation}
\label{sec:efficiency_challenge}

Frequent online update of fast weights is inefficient due to memory bandwidth limitations. Consequently, previous works~\cite{sun2023retentive, gu2023mamba, yang2024gated, qinvarious, yangparallelizing} often employ customized kernels that keep fast weights in SRAM across updates to reduce memory load. 
However, this strategy typically requires fast weights to evolve mostly independently within SMs to reduce communications, which is not valid for large nonlinear states (e.g., the nonlinear SwiGLU fast weight in Sect.~\ref{sec:ttt_layer} and the Muon update in Sec.~\ref{sec:update_rule}).   Moreover, developing such kernel code is cumbersome, with far longer development cycles than native PyTorch code, hindering rapid research exploration.

\begin{figure}[t!]
    \centering
    \includegraphics[width=\textwidth]{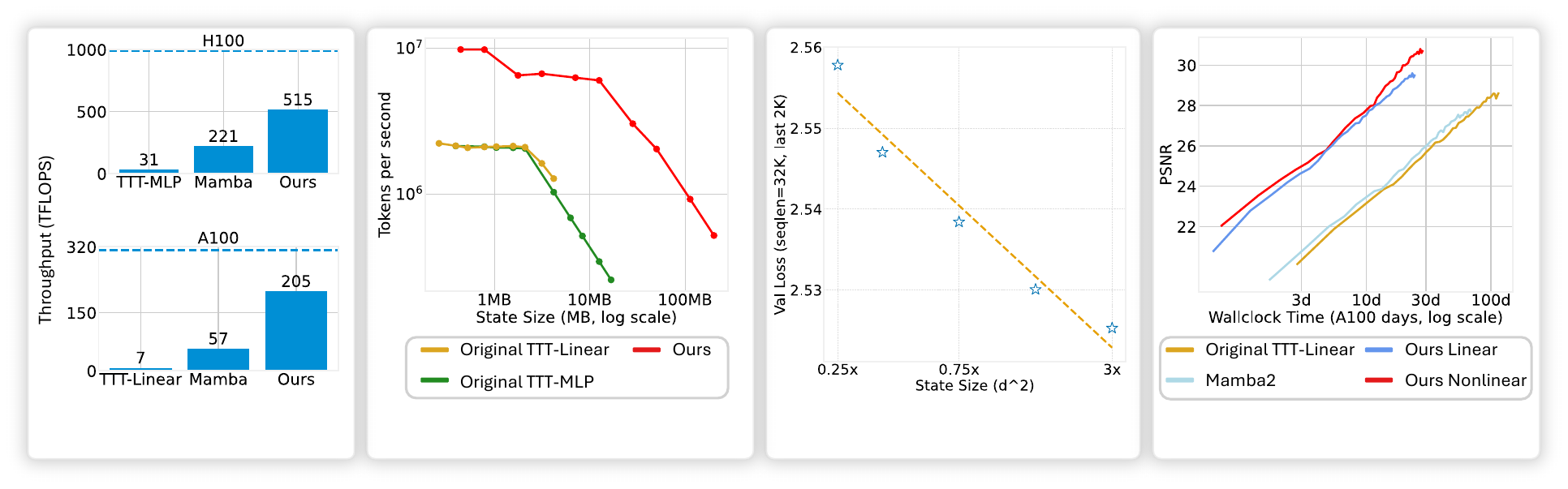}
    \put(-388,12){\capfont \textbf{(a)} GPU Throughput}
    \put(-300,12){\capfont \textbf{(b)} Efficient state scaling}
    \put(-202,12){\capfont \textbf{(c)} Effective state scaling}
    \put(-96,12){\capfont \textbf{(d)} Training Efficiency}
    \vspace{-0.2in}
    \caption{Using larger chunk sizes significantly improves GPU utilization compared to the original test-time training (TTT) method that even uses customized kernels \textbf{(a)}. This enhanced utilization enables efficient and effective scaling to larger state sizes \textbf{(b)}, \textbf{(c)}, leading to better overall performance in less wall-clock time (\textbf{d}). The dotted line in \textbf{(a)} is the theoretical peak BF16 throughput of the GPU. Panel \textbf{(c)} measure average validation loss of the last 2K tokens in sequences processed by a \methodname{} language model across varying state sizes, demonstrating benefits of larger state size. Panel \textbf{(d)} compares performance versus training time across different baselines on the novel view synthesis benchmark. Further experimental details can be found in Sec.~\ref{sec:appen_figure_1_details}.
    \vspace{-0.1in}
    }
    \label{fig:compare_with_ttt}
\end{figure}
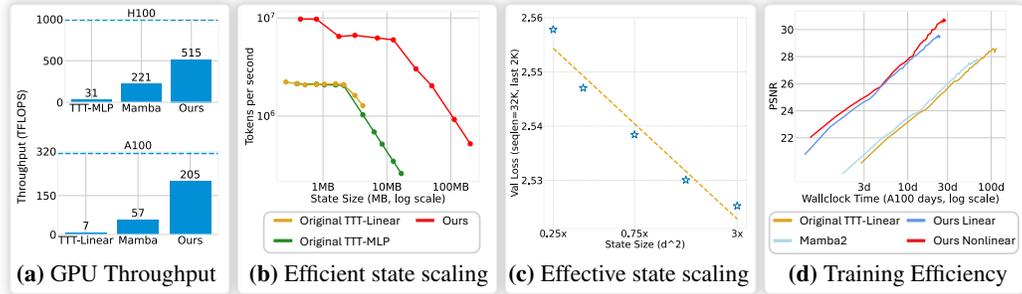

On the other hand, a PyTorch-based implementation, while simpler, is typically bounded by memory speed. As an illustration, consider a PyTorch implementation of simple MLP fast weight, the core of which is a matrix multiplication between fast weight (e.g., $h \times h$ matrix) and the mini-batch input ($b \times h$ where b is the chunk size).
The ideal compute-to-memory ratio is:
\begin{align} \label{eq:comp_to_mem_ratio}
r&= \frac{2 h^2 b}{2h^2+ 4 hb}=\frac{h / 2}{1 + \frac{h}{2b}} = \frac{b} {1  + \frac{2b}{h}} \le \min(h / 2, b).
\end{align}
Here, $2 h^2 b$ is the FLOPs to for matrix multiplication, the denominator $2h^2+ 4 hb$ is the memory workload for two input matrices and the output in BF16 (2 bytes).
Small fast weight size (e.g., $h=64$) or small chunk size (e.g., $b=16$) will bound the ratio $r$ far below the theoretical peak (e.g., $290$ FLOPs per byte on H100), making the operation memory-bound and limiting compute usage.

In light of this, we advocate for using large chunk sizes (from $2048$ to $1$M).
This allows us to achieve higher throughput (Fig.~\ref{fig:compare_with_ttt}a) leading to better performance in less training wall-clock time(Fig.~\ref{fig:compare_with_ttt}d).
Our design also allows the state size to be  
scaled up efficiently(Fig.~\ref{fig:compare_with_ttt}b), leading to significant results improvement with such scaling 
(Fig~\ref{fig:compare_with_ttt}c, Fig.~\ref{fig:exp-state-optimizer}a).
Our architecture achieves a state-to-parameter size ratio $\ge\!40\%$, which is an order of magnitude 
larger than previous methods' ratio of $0.1\%$ to $5\%$. Detailed pseudocode is provided in Appendix~\ref{alg:full_lact_layer}.

\textbf{Parallelism over the sequence length dimension}, in addition to the batch and head dimensions, is crucial to achieve high occupancy when handling long sequences (where the batch size is often small). Linear Attention variants like Mamba~\cite{gu2023mamba}, Gated Linear Attention~\cite{yang2024gated} and DeltaNet~\cite{yangparallelizing} enable such parallelism by utilizing the associative property of linear recurrence. Attention~\cite{vaswani2017attention,dao2023flashattention} can be parallelized along the sequence length dimension using online softmax~\cite{milakov2018online}, a key improvement in FlashAttention-2~\cite{dao2023flashattention} over FlashAttention-1~\cite{dao2022flashattention}. For test-time training with non-linear updates, sequence dimension parallelism can only be implemented within online chunks, further motivating the use of extremely large chunk sizes. When implementing large-chunk TTT with PyTorch, this sequence dimension parallelism within a device across multiple thread blocks is automatically handled by PyTorch and low-level compilers. An example of such sequence parallelism across multiple devices is provided in Section~\ref{sec:parallelism}, with pseudocode in Appendix~\ref{alg:lact_chunk_cp}.

\section{\methodname{} Model Architecture} \label{sec:model}

\begin{figure}[t!]
    \centering
    \includegraphics[width=1.0\textwidth]{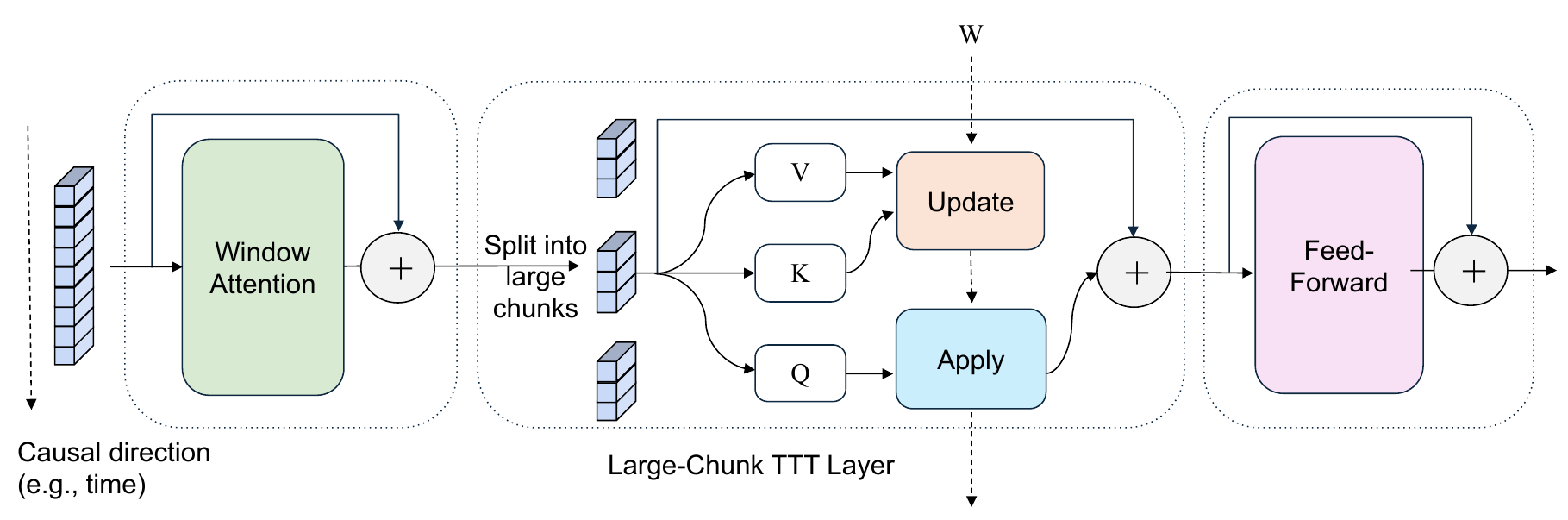}
    \vspace{-0.20in}
    \caption{The basic diagram for a \methodname{} block. 
    The large-chunk TTT layer updates the fast weight $W$ to store historical context information,
    while the window attention handles the locality and internal structures within the chunk. 
    The solid line denotes the information flow over model depth and the dashed line denotes the information flow over time (i.e., the fast weight $W$ passing through chunks).
    Various instantiations in Sec.~\ref{sec:method} use different chunk sizes and window attention types according to the specific data structure. Additionally, window attention and large-chunk TTT layers can be combined within the same layer by sharing the QKV and summing their outputs; this in-layer mixing is used in our language modeling and video generation experiments (see Appendix~\ref{alg:lact_hybrid} for such pseudocode).}
    \vspace{-0.10in}
    \label{fig:lact_model_arch}
\end{figure}

As shown in Fig.~\ref{fig:lact_model_arch},
\methodname{} block consists of three types of layers: a window attention layer, a large-chunk TTT layer, and a feed-forward layer.
Each layer is equipped with residual connections~\cite{he2015deep} following the practice in Transformer~\cite{vaswani2017attention}.
The window attention layer performs local self-attention 
to capture the local dependency.
In the TTT layer, we split the sequence into large chunks.
The history context is gradually compressed into the fast weights through an `update' operation (regarding key vectors $K$ and value $V$), and latest weight is `applied' to the current query vector (Q) for computing its corresponding output.
The feed-forward layer performs channel mixing as in Transformer.
We omit several linear and normalization layers in Fig.~\ref{fig:lact_model_arch} for clarity and details are in Appendix~\ref{sec:ttt_pseudocode}.
Our framework offer great flexibility in handling diverse data types. 
In this section, we 
present the general designs in our approach and later  describe data-specific variations 
in Sec.~\ref{sec:method}.

\subsection{Large-Chunk TTT Layer}
\label{sec:ttt_layer}
Different from the per-token update in Eqn.~\ref{eq:ttt_update}, the chunk-wise update computes the gradient of the summed loss over all keys $\{k_i\}$ and values $\{v_i\}$ within the chunk.
As the chunk size is large, weight updates are performed infrequently. This enables more sophisticated weight-update rule designs (discussed in Sec.~\ref{sec:update_rule}) and amortizes the update cost.
The `update' operation for the fast weight is:
\begin{align}
g &=  \nabla_{W}  \sum_{i=1} ^b \eta_i \mathcal{L}  \big(f_W(k_i), v_i\big) \label{eq:ttt_chunk_gradient} \\
W &\leftarrow \mathrm{weight\mbox{-}update} (W, g),
\label{eq:ttt_chunk_update}
\end{align}
where $b$ is the chunk size, $g$ is the gradient of the fast-weight loss function, 
and $\eta_i$ is the learning rate of each token (usually predicted from input tokens).
The `apply' operation $o_i = f_W(q_i)$ is the same as Eqn.~\ref{eq:ttt_apply} and 
 all query vectors $\{q_i\}$ in the chunk share the same updated fast weight $W$.

Motivated by recent 
LLMs~\cite{touvron2023llama}, we adopt SwiGLU-MLP~\cite{shazeer2020glu} without bias terms as the fast-weight 
network. 
Our fast weights consists of three weight matrix $W = \{W_1, W_2, W_3\}$, and the network 
is:
\begin{equation}
f_W(x) = W_2 \left[ \mathrm{SiLU}(W_1 x) \circ (W_3 x) \right]
\label{eq:swiglu}
\end{equation}
where $\circ$ is an elementwise multiplication. 
We apply a simple dot product loss as our loss function:
\begin{align}
    \mathcal{L} \big(f_W(k_i), v_i\big) = -f_W(k_i)^\top v_i 
\label{eq:loss}
\end{align}

\begin{figure}[t!]
    \centering
    \includegraphics[width=1.0\textwidth]{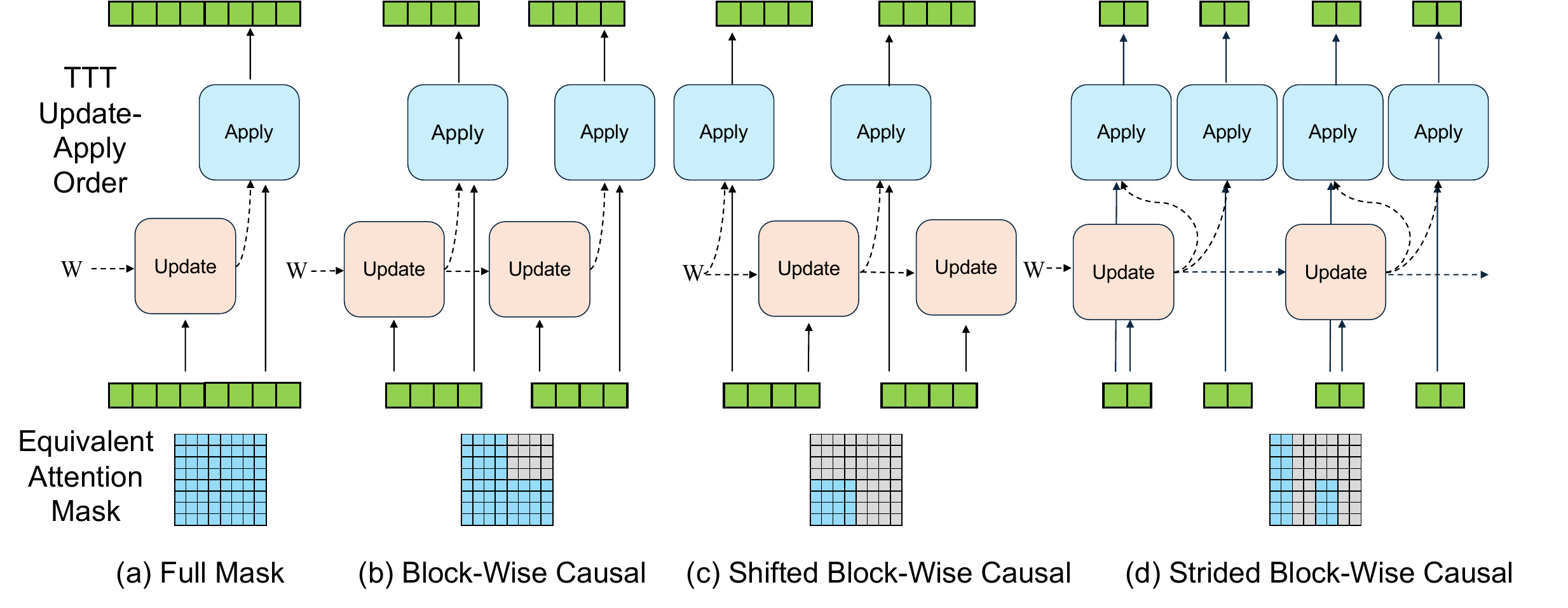}
    \vspace{-0.3in}
    \caption{Different `Update' and `Apply' orders and their
    equivalent attention mask.  
    A blue mask in i-th row and j-th column means the i-th token's output depends on the j-th token.
    }
    \vspace{-0.15in}
    \label{fig:large_chunk_recurrence}
\end{figure}

\noindent\textbf{Execution orders for `apply' and `update'.
}
Note that the `update' operation and `apply' operation of TTT are decoupled,  
and we can set the chunk size adaptively and apply these operation in different 
orders; this allows us to model diverse kinds of data dependencies, similar to different 
attention masks in self-attention. 
Figure~\ref{fig:large_chunk_recurrence} illustrates this concept.
In Figure~\ref{fig:large_chunk_recurrence}a, when the chunk size equals the full sequence length, performing the apply followed by the update operation is conceptually similar to full attention. 
Using update and apply alternately leads to 
a block-wise causal mask (Fig.~\ref{fig:large_chunk_recurrence}b), where the block size corresponds to the chunk size.
Switching the order between the two operations results in the a shift in the mask (Fig.~\ref{fig:large_chunk_recurrence}c). 
This shifted mask 
does not leak future information within the chunk and is important when building the full causal mask 
in Language Modeling (Sec.~\ref{sec:method-lm}).
Moreover, only updating on a subset of chunks 
and applying to all
(Figure~\ref{fig:large_chunk_recurrence}d) is analogous to strided 
block-wise causal mask.


\subsection{Non-Linear Update of Fast-Weight }
\label{sec:update_rule}

Fast-weight updates in TTT repeatedly accumulate gradients, and thus suffer from magnitude explosion or decayed memory.
Large-chunk TTT allows non-linear updates to improve stability and effectiveness while preserving efficiency.
For the `weight\mbox{-}update' operation in Eqn.~\ref{eq:ttt_chunk_update}, our vanilla implementation involves gradient descent  followed by weight normalization:
\begin{align}
\mathrm{weight\mbox{-}update} (W, g) & = \mathrm{L2\mbox{-}Normalize} (W - g)  .
\label{eq:fast_weight_normalization_nomuon}
\end{align}
We have also explored a more robust nonlinear Muon~\cite{jordan2024muon} update rule~\footnote{Muon requires weights in matrix form, and our current fast-weight function SwiGLU-MLP has three matrices as the weights (i.e., $W_1\mbox{,}W_2\mbox{,}W_3$ in Eqn.~\ref{eq:swiglu}). } with weight normalization:
\begin{align}
\mathrm{weight\mbox{-}update} (W, g) & = \mathrm{L2\mbox{-}Normalize} (W - \mathrm{Muon}(g))  
\label{eq:fast_weight_normalization}
\end{align}

\noindent\textbf{Fast-weight normalization.} 
We apply L2 weight normalization~\cite{salimans2016weight} to the updated fast weights along the input dimension. We do not use explicit weight-decay term as in previous methods~\cite{behrouz2024titans, dao2024transformers, yang2024gated, sun2023retentive}. 
When the network is conceptually rotated 90 degrees, treating the sequence dimension as the depth of a virtual model, the test-time training updates act as residuals over time~\cite{he2015deep}. In this view, our fast-weight normalization is analogous to the \textit{post-layer norm} in Transformer architectures, which constrains activation scales within the residual path.

\noindent\textbf{Muon-update rule.} 
Essentially, Muon normalizes the spectral norm of matrix gradient using Newton-Schulz iterations.
In short, let $g = U S V^T$ be the Singular Value Decomposition(SVD) of the gradient $g$, then Muon operator approximately converts the gradient as:
\begin{align}
    \mathrm{Muon}(g) &\simeq UV^T
\end{align}
Muon also improves the numerical stability in our setup. 
For example, 
the learning rate  ($\eta_i$ in Eqn.~\ref{eq:ttt_chunk_gradient}) now only reflects the relative importance of tokens within a chunk as Muon normalizes the absolute scale. See \cite{jordan2024muon} and Appendix~\ref{sec:appen_lact_details} for analysis of its computational cost. 



\subsection{Window Attention}
The large-chunk TTT layer treats data as sequences of sets because its fast weight updates inherently disregard token order and spatial locality within each chunk.
However, many data modalities—such as videos (sequences of grids), image collections (sets of grids), or text (1D sequences)—do not fully align with this set-based perspective .
For these modalities, intra-chunk structure and locality are vital for capturing the overall data structure. We therefore integrate local window attention (either causal or bidirectional) alongside TTT layers to handle data structure within a chunk. Moreover, window attention efficiently handles localities in the data, enabling the TTT layer to focus its fixed-size fast weight capacity on modeling non-local dependencies. This hybrid strategy is also employed in other notable works like BASED~\cite{arora2024simple}, GAU~\cite{hua2022transformer} and InifinitAttention~\cite{munkhdalai2024leavecontextbehindefficient}. 
In summary, \methodname{} is a hybrid architecture with the quadratic-compute attention for local structure and linear-compute TTT for non-local context.

\subsection{Context Parallelism}
\label{sec:parallelism}

Context Parallelism (CP) partitions the sequence along the context length dimension and distributes the shards across multiple devices for parallel computing.
The feed-forward layer and window attention are local operators thus natively support CP.
For TTT layer, small chunks hardly support CP thus tensor parallelism (i.e., parallel over the heads) is preferred.
Our large-chunk TTT layer allows CP by sharding the tokens within a chunk.
Suppose each shard contains $s$ tokens,
the fast weight gradient of the chunk is the sum over all shard's gradients given the linearity of the gradients:
\begin{align}
g &=  \nabla_{W}  \sum_{j=1}^{\text{shards}} \sum_{i=1}^{\text{s}} \eta_i \mathcal{L}_i  = \sum_{j=1}^{\text{shards}} \nabla_{W}   \sum_{i=1}^{\text{s}} \eta_i \mathcal{L}_i
\label{eq:context_parallelism}
\end{align}
This can be implemented through distributed all-reduce-sum and is 
logically the same as Distributed Data Parallelism (DDP), except that the parameters are the fast weights and input data are the tokens in the chunk.
We adopt such parallelism in training the novel view synthesis task (see Sec.~\ref{sec:method-nvs}) 
and observe minimal throughput overheads (1\% to 3\%).
\methodname{} architecture is compatible with other parallelism strategies (e.g., data parallelism, pipeline parallelism, and tensor parallelism). See Appendix for pseudocode on implementing context parallelism(Alg.~\ref{alg:lact_chunk_cp}) and tensor parallelism(Alg.~\ref{alg:lact_tp}) for \methodname{}.

\section{\methodname{} for N-Dimensional Data} \label{sec:method}

In this section, we introduce the three tasks we address using \methodname{}---novel view synthesis, language modeling, and autoregressive video generation.
These tasks have different inherent data structures and we address them with corresponding design choices.
The full model architecture details for these data types are provided in Appendix~\ref{sec:appen_model_arch}.

\subsection{Novel View Synthesis - Image Set}
\label{sec:method-nvs}
Novel view synthesis (NVS)\cite{mcmillan2023plenoptic, levoy2023light} aims to render images of a static scene from previously unseen viewpoints. Formally, 
given a set of $N$ input posed images $\{ (  \img_i, P_i ) \}_{i=1}^N$ of a static scene, 
where $I_i \in \mathbb{R}^{H\times W \times 3}$ is an RGB image and $P_i$ is its corresponding camera pose,
the model needs to synthesize new images from novel camera poses that typically do not overlap with the input views.

We find that NVS is an effective test bench 
for evaluating a model's online memory and compression capabilities. 
Firstly, NVS is
challenging as it requires spatial compression, dense retrieval, and basic physical reasoning.
Secondly, NVS can be formulated as a non-generative task, significantly reducing training computation and the need for extensive model parameters to store world knowledge, thereby enabling rapid experimentation.
Thirdly, the substantial redundant information in dense input views incentivizes the model to learn effective compressions.
Given these observations, we use NVS for our initial research iterations. We find that some of the insights gained are transferrable to other tasks.



Our NVS model follows the basic \methodname{} diagram in Sec.~\ref{sec:model}.
Both the posed input images and poses of the target novel views are tokenized by patchify and linear layers, following LVSM~\cite{jin2024lvsm}.
The window attention exactly covers the tokens from a single image.
The \methodname{} layer adapts a single-round of strided block-wise causal mask (Fig.~\ref{fig:large_chunk_recurrence}d), which updates the fast weight using all input image tokens, and applies to both the input and target tokens. The \textit{update} step resembles a prefill stage,  while the \textit{apply} operation resembles parallel decoding. During rendering of novel views, each test-time training layer functions as a static weight layer, making the entire model a static vision transformer~\cite{dosovitskiy2020image}. We illustrate this design in Figure~\ref{fig:appen_lvsm_model}.

\subsection{Language Modeling - Text Sequence}
\label{sec:method-lm}

 Autoregressive language models predict the probability distribution of the next token given preceding tokens, $p_{\theta}(x_n|x_1, \dots, x_{n-1})$.  Text sequences lack inherent chunk structures, so for \methodname{}, we define chunk size as a hyperparameter (e.g., 2048 or 4096 tokens).  We utilize the shifted block-wise causal mask as in Fig.~\ref{fig:large_chunk_recurrence}(c) for the TTT apply-update sequence  to avoid seeing future tokens in a chunk.
 Since \methodname{} lacks per-token causality within each chunk, we employ sliding window attention—with window size equal to the chunk size—to efficiently model per-token causal dependencies. The sliding window is integrated into the same TTT layer with shared QKV similar to GAU~\cite{hua2022transformer}. We illustrate the detailed architecture in Fig.~\ref{fig:appen_lm_model} and pseudocode~\ref{alg:lact_hybrid}.

\subsection{Autoregressive Video Diffusion - Image Sequences}

\label{sec:method-arvideo}


Chunkwise autoregressive video diffusion iteratively denoises a number of subsequent video frames, conditioned on the previously generated clean frames, where each chunk can contain thousands of visual tokens. 
We use teacher-forcing training by interleaving noisy and clean frame chunks.
Specifically, a video of N frame chunks is structured as:
\begin{equation}
    S = [X_1^{\text{noise}}, X_1, X_2^{\text{noise}}, X_2, \ldots, X_N^{\text{noise}}]
    \label{eq:video_sequence_basic}
\end{equation}
where each noisy chunk $X_i^{\text{noise}}$ is produced by adding unit Gaussian noise $\epsilon$ to the $i$-th clean video chunk as $X_i^{\text{noise}} = X_i (1 - t_i) + \epsilon t_i$ and $t_i \in [0, 1]$ denotes the strength of chunk-independent noise. 

To handle such a data structure, we employ the strided block-wise causal mask in Fig.~\ref{fig:large_chunk_recurrence}d for  \methodname{}.
Specifically, it \textit{applies} fast weights to each chunk sequentially while only \textit{updating} fast weights on clean chunks.
This simple strategy ensures that each denoising operation only accesses previously cleaned frames. 
The windowed attention uses a non-overlapping window with 2 consecutive chunks (i.e., $[X_i, X_{i+1}^\text{noise}]$) to build temporal and spatial locality.
Within each window, the attention from $X_i$ to  $X_{i+1}^\text{noise}$ is excluded.
We incorporate the first noisy chunk by shifting all attention and TTT masking patterns similar to Fig.~\ref{fig:large_chunk_recurrence}c.
The details of this hybrid architecture and more efficient trainings are in the Appendix~\ref{sec:appen_video_model_arch}.

\section{Experiments} \label{sec:experiments}
\vspace{-0.3em}
In this section, we present our experiment results on novel view synthesis (Sec.~\ref{sec:exp_nvs}),
language modeling (Sec.\ref{sec:exp_language}), and autoregressive video generationo (Sec.~\ref{sec:exp_video}), and an in-depth analysis (Sec.~\ref{sec:analysis}) of different design choices.
Tab.~\ref{tab:experiment-summary} summarizes key factors in each experiment. 
When comparing with linear-cost baselines, we augmented them with the same window attention for fair comparisons. 
The full experimental details for all tasks are provided in Appendix~\ref{sec:appen_experiment_details}.
\vspace{-0.3em}

\begin{table}[t!]
  \caption{Summary of our experiments on three different data structures. `d' denotes model dimension. The state size denotes the size of the fast weight per model block.}
  \centering
  \scriptsize
  \resizebox{\textwidth}{!}{%
    \begin{tabular}{@{}%
        l  
        l  
        l  
        l  
        l  
        l  
        l  
        l  
        l  
      @{}}
      \toprule
      Task name & Data Structure & Chunk Size 
      & State Size 
        & Model Size & Max Length 
        & Context Parallelism \\
      \midrule
      Novel View Synthesis
        & Image set
        & Full sequence
        & $6d^2$
        & 0.3B
        & 1M
        & Within-chunk parallel
         \\

      AR Video Diffusion
        & Image sequence
        & Three frames
        & $3d^2$, $0.75d^2$
        & 1.3B, 14B
        & 56160
        & Head-dim parallel
          \\

      Language Models
        & 1D Sequence
        & 2K, 4K tokens
        & $0.75d^2$
        & 0.7B, 3B
        & 32768
        & N/A
         \\
      \bottomrule
    \end{tabular}%
  }
  \vspace{-0.2in}
  \label{tab:experiment-summary}
\end{table}

\subsection{Novel View Synthesis}\label{sec:exp_nvs}

\noindent\textbf{Datasets \& metric.} We evaluate our approach on 
both object-level and scene-level datasets.
We use Objaverse dataset \cite{deitke2023objaverse} for object-level training, 
following the setup from LVSM \cite{jin2024lvsm} and GS-LRM \cite{zhang2024gs}.  
After training, we perform evaluations on the Google Scanned Objects (GSO) dataset \cite{downs2022google}, at resolutions of $256\times256$ and $512\times512$. Each evaluation involves 4–48 input views and 8 novel views per object.
For scene-level evaluations, we adopt the challenging DL3DV scene dataset \cite{ling2024dl3dv}, with over 11K training scenes and 140 testing scenes, each with approximately 300 views.
Evaluations are at a resolution of $960\times536$.
Performance is measured by Peak Signal-to-Noise Ratio (PSNR) at novel views, with additional metrics provided in the Appendix~\ref{sec:appen_nvs_details}.

\noindent\textbf{Model details.}
Each block of model
has a per-image window attention layer,  a SwiGLU-MLP large-chunk TTT layer, 
and a feed-forward layer. 
The default model totals 312M parameters, including 84M fast weights ($6 d^2$ per block). 


\noindent\textbf{Baselines.}
For object-level evaluation, we use two baselines: a full-attention model and a Perceiver-style register-attention model~\cite{jaegle2021perceiver}. The full-attention baseline replaces TTT layers with block-wise causal attention layers, enabling bidirectional interaction among input tokens and cross-attention from novel views. The Perceiver-style baseline compresses input tokens into 4096 registers, decoding novel views via cross-attention to these registers.
For scene-level evaluation, we compare with LongLRM~\cite{ziwen2024long}, a state-of-the-art model combining Mamba~\cite{gu2023mamba} and full attention for 3D Gaussian splat predictions, as well as pure optimization-based 3D Gaussian splatting methods. Table~\ref{tab:attention_comparison} summarizes the computational complexities of all models.

\noindent\textbf{Training details.}
For object dataset, we train all models with $1.25$ trillion tokens with progressive resolutions.  For scene dataset, we train our model with $1.8$ trillion tokens with progressively higher resolutions and more views, at a maximal sequence length of 1 million tokens.  High-resolution models are trained with inner-chunk context parallelism (Sec.~\ref{sec:parallelism}). See more details in Sec.~\ref{sec:appen_nvs_details}.  

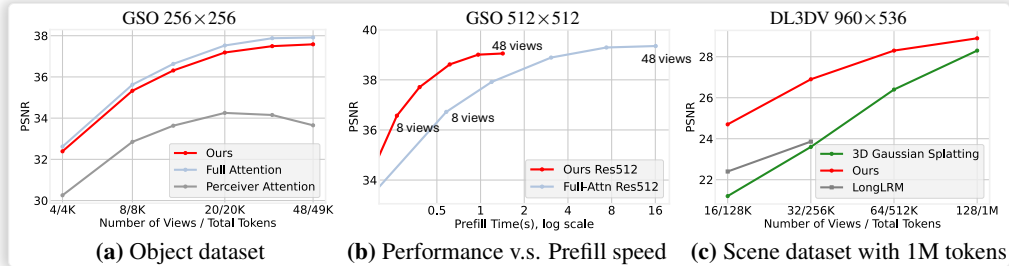
\begin{figure}[t!]
    \vspace{-0.1in}
    \centering
    \includegraphics[width=\textwidth]{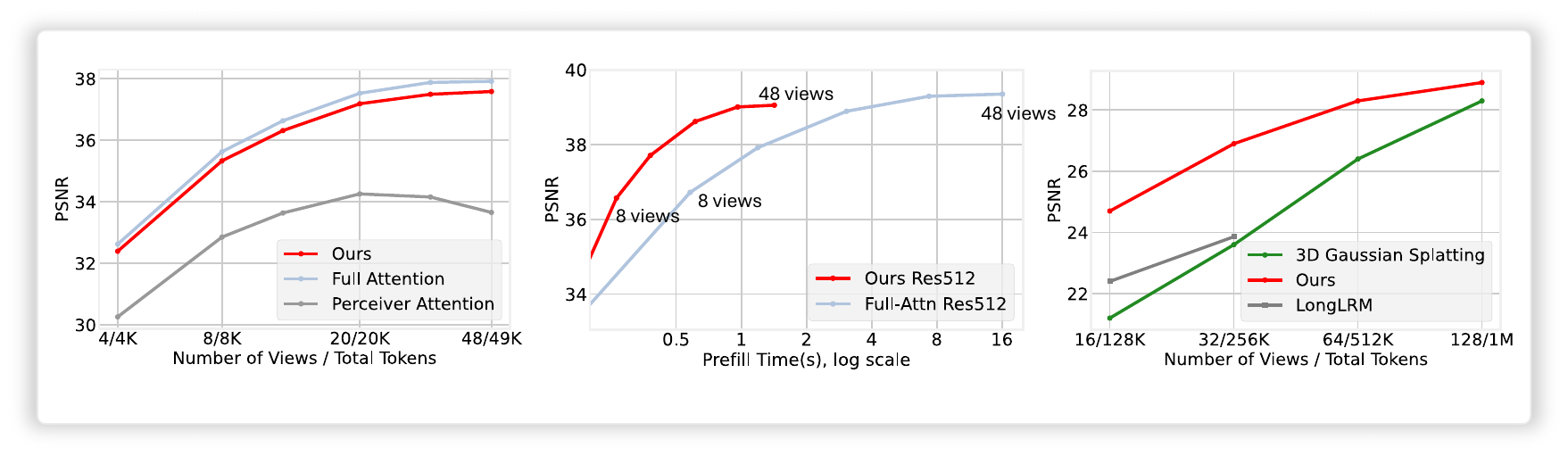}
    \put(-345, 100){\capfontscriptsize GSO 256$\times$256}
    \put(-215, 100){\capfontscriptsize GSO 512$\times$512}
    \put(-100, 100){\capfontscriptsize DL3DV 960$\times$536}
    \put(-355,11){\capfont  \textbf{(a)} Object dataset }
    \put(-260,11){\capfont \textbf{(b)} Performance v.s. Prefill speed}
    \put(-130,11){\capfont \textbf{(c)} Scene dataset with 1M tokens}
    \vspace{-0.15in}
    \caption{\textbf{(a, b)} our method achieves quality comparable to full-attention models with significantly lower prefill latency,  and it clearly outperforms perceiver-attention baselines. \textbf{(c)} On the high resolution scene dataset, our approach surpasses LongLRM, limited to 32 views, and outperforms 3D Gaussian Splatting with sparse views, remaining competitive up to 128 input views (1M total tokens). }
    \vspace{-0.14in}
    \label{fig:view_syn_results}
\end{figure}

\noindent\textbf{Results.}
Experimental results and analysis are presented in Figure~\ref{fig:view_syn_results}.

\begin{table*}[t!]
\centering
\vspace{-0.03in}
\caption{Complexities of methods on novel view synthesis w/ $n$ input.
Prefill and rendering speed are measured on A100 with 48 512$\times$512 input images
(196K input tokens, 4K decoding tokens).}
\label{tab:attention_comparison}
\resizebox{\textwidth}{!}{%
\begin{tabular}{lccccll}
\toprule
 & \textbf{State Size} & \textbf{Prefill Compute} & \textbf{Decoding Compute} & \textbf{\# Params} & \textbf{Prefill speed} & \textbf{Rendering FPS} \\
\midrule
Full attention & $O(n)$ & $O(n^2)$ & $O(n)$ & 284M & 16.1 s & 2.3 FPS \\
Perceiver Attention & $O(1)$ & $O(n^2)$ & $O(1)$ & 287M & 16.8 s & 34.4 FPS \\
Ours & $O(1)$ & $O(n)$ & $O(1)$ & 312M & 1.4 s & 38.7 FPS \\
\bottomrule
\end{tabular}%
}
\vspace{-0.1in}
\end{table*}

\subsection{Language Modeling}\label{sec:exp_language}

\noindent\textbf{Datasets \& Metrics.}
We train our models on the Long-Data-Collections dataset~\cite{long_data_collection}, using approximately 60B tokens from its total 68.8B tokens. For evaluation, we employ the per-token loss metric from~\cite{linforgetting}, assessing models' ability to effectively use the full context. A monotonically decreasing loss indicates successful context utilization, whereas plateauing suggests limited context usage. Additionally, we report retrieval accuracy~\cite{hsiehruler} at various sequence lengths.


\noindent\textbf{Model details.} We remove the window-attention layer from the original the \methodname{} block, integrating a sliding window-attention(SWA) layer directly into the Large-Chunk TTT layer. Following GAU~\cite{hua2022transformer}, SWA shares Q, K, and V vectors with the fast-weight network, with additional per-channel scaling and shifting on Q and K. 
The pseudocode for this design is in Algorithm~\ref{alg:lact_hybrid}.

\noindent\textbf{Baselines.}
We compare against full attention, Gated Linear Attention (GLA)~\cite{yang2024gated}, DeltaNet~\cite{schlag2021linear,yangparallelizing}. To ensure fairness, we enhance both GLA and DeltaNet with the same sliding window attention. Based on prior work~\cite{linforgetting, xiong2023effective, men2024base} highlighting the importance of a large RoPE~\cite{su2023roformer} base for long-context transformer training,  we adopt a RoPE base of 1 million for training with 32K token contexts.
Tab.~\ref{tab:LM_baseline} summarize the mechanism and training throughput of all methods. 

\begin{table*}[t!]
\centering
\caption{Comparison of baseline methods in terms of state size, training throughput (measured in tokens per second, TPS), update rules, and memory read-out mechanisms. Training throughput is evaluated using a 3B-parameter model with 32K-sequence length on A100-40GB GPUs.}
\label{tab:LM_baseline}
\resizebox{\textwidth}{!}{%
\begin{tabular}{llllll}
\toprule
 & \textbf{State size}  & \textbf{Train TPS} &  \textbf{Update Rule} & \textbf{Memory read-out} \\
\midrule
Transformer     & –       & 4.1K     & – & – \\
Transformer SWA & –       & 6.4K   & – & – \\
\midrule
\multicolumn{5}{l}{ \textit{Per-token recurrence} } \\
GLA SWA         & $384d$  & 5.0K   &
$\displaystyle \mathbf{S}_t \leftarrow  \mathbf{S}_{t-1}\mathrm{Diag}(\boldsymbol{\alpha}_t) + \mathbf{v}_t \mathbf{k}_t^\top$ 
& $\displaystyle \mathbf{o}_t = \mathbf{S}_t \mathbf{q}_t$ \\
DeltaNet SWA    & $128d$  & 5.1K   &
$\displaystyle \mathbf{S}_t \leftarrow  \mathbf{S}_{t-1}(\mathbf{I} - \beta_t \mathbf{k}_t \mathbf{k}_t^\top) + \beta_t \mathbf{v}_t \mathbf{k}_t^\top$ 
& $\displaystyle \mathbf{o}_t = \mathbf{S}_t \mathbf{q}_t$ \\
\midrule
\multicolumn{5}{l}{ \textit{Large-chunk recurrence} } \\
Ours GD       & $2304d$ & 5.0K    
& $ W \leftarrow \mathrm{L2norm}(W - \sum_i^b \eta_i \nabla_W \mathcal{L}_i)$
& $\displaystyle \mathbf{o}_t = f_W(\mathbf{q}_t)$  \\
Ours  Momentum       & $2304d$ & 4.9K    
& $M \leftarrow \beta M + \sum_i^b \eta_i \nabla_W \mathcal{L}_i;\; W \leftarrow  \mathrm{L2norm}(W - M)$
& $\displaystyle \mathbf{o}_t = f_W(\mathbf{q}_t)$  \\
Ours Muon       & $2304d$ & 4.3K  
& $ M \leftarrow  \beta M + \sum_i^b \eta_i \nabla_W \mathcal{L}_i;\;  W \leftarrow  \mathrm{L2norm}(W - \mathrm{Muon}(M))$
& $\displaystyle \mathbf{o}_t = f_W(\mathbf{q}_t)$ \\
\bottomrule
\end{tabular}%
}
\end{table*}

\noindent\textbf{Training details.} We trained models at two scales using a 32768-token sequence length: a 760M-parameter model trained for 40B tokens with a 2048-token sliding window, and a 3B-parameter model trained for 60B tokens with a 4096-token sliding window. See more details in Sec.~\ref{sec:appen_exp_language}.  

\begin{figure}[t!]
    \centering
    \vspace{-0.1in}
    \includegraphics[width=\textwidth]{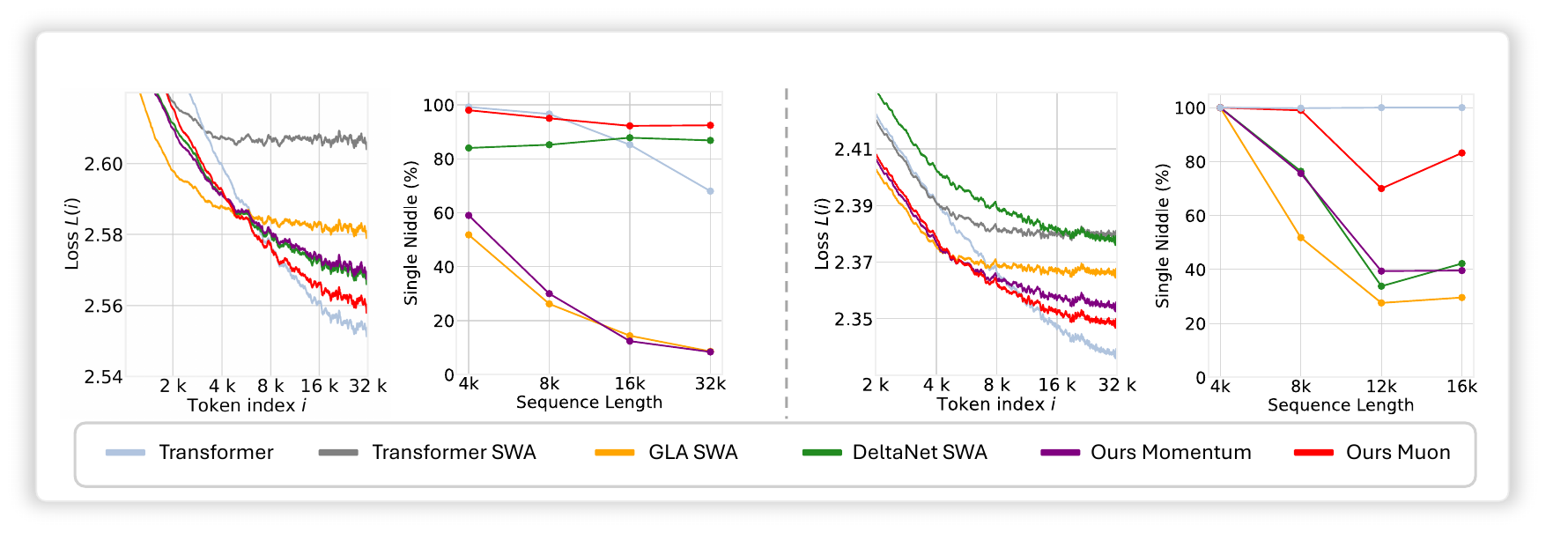}
    \put(-388, 116){\capfont \textbf{(a)} 760M Validation loss}
    \put(-288, 116){\capfont \textbf{(b)} 760M S-NIAH-1}
    \put(-190, 116){\capfont \textbf{(c)} 3B Validaiton loss}
    \put(-94, 116){\capfont \textbf{(d)} 3B S-NIAH-2}
    \vspace{-0.1in}
    \caption{Language Model results. 
    \textbf{(a, c)} Our model achieves lower per-position loss at larger token indices compared to GLA and DeltaNet at both 760M and 3B scale, indicating stronger long-context modeling capability.
\textbf{(b, d)} Our model consistently outperforms GLA and DeltaNet in retrieval accuracy. Furthermore, our Muon variant consistently outperforms our Momentum variant.
    }
    \label{fig:lm_results}
\end{figure}
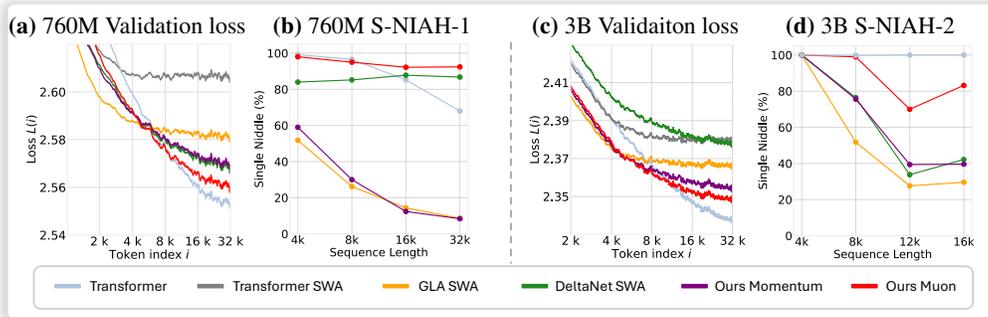

\noindent\textbf{Results.} Please refer to Fig.~\ref{fig:lm_results} for experiment results and analysis. See more results in Sec.~\ref{sec:appen_exp_language}.

\subsection{Autoregressive Video Diffusion}\label{sec:exp_video}
We fine-tune the pretrained Wan 2.1~\cite{wang2025wan} text-to-video diffusion model into an autoregressive video diffusion model. Specifically, we replace all bidirectional attention layers with our \methodname{} layers combined with sliding window attention. The sliding window attention uses a window size spanning two autoregressive chunks.  


\noindent\textbf{Datasets.} We fine-tune the model using an internal, filtered proprietary collection of videos, each accompanied by a short text prompt generated by a visual language model.

\noindent\textbf{Training details.}
Following~\cite{esser2403scaling, wang2025wan}, we employ time-step shifting and denoising loss weighting using a logit-normal distribution. we train on 5-second videos at 16 FPS and 480$\times$832 resolution, autoregressively denoising in 3 latent-frame chunks. Later we fine-tune the 1.3 billion parameter model with 10 second videos and 14 billion parameter model with 8.8 second videos. Each 8.8-second clip contains 56,160 visual tokens, resulting in interleaved noisy-clean chunks totaling 107K tokens under teacher-forcing training. We use sequence parallelism for MLP layers and tensor parallelism (sharding heads across devices) for TTT and window attention layers. Full details are listed in Sec.~\ref{sec:appen_exp_video}.


\noindent\textbf{Baselines.}
We compare our method against three baselines: sliding window attention (SWA) alone, Mamba2~\cite{dao2024transformers} combined with SWA (using a similar parallel combination strategy as our method), and full block-wise causal attention.  

\begin{figure}[t!]
    \centering
    \includegraphics[width=\textwidth]{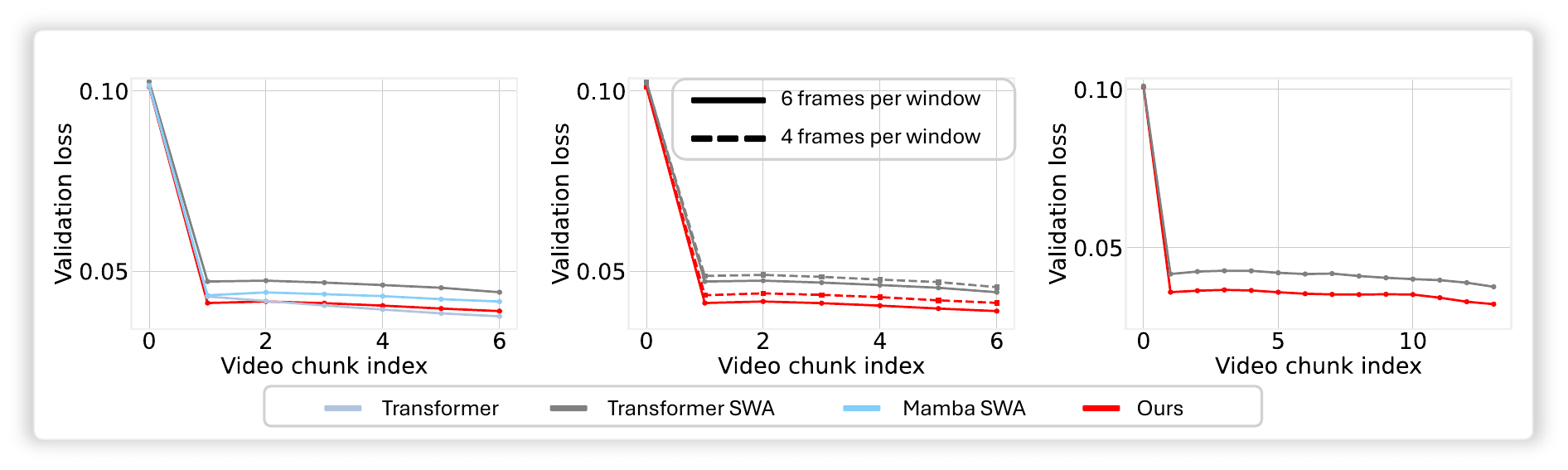}
    \put(-376, 103){\capfont \textbf{(a)} Comparison with baselines}
    \put(-245, 103){\capfont \textbf{(b)} Ablate on window size}
    \put(-110, 103){\capfont \textbf{(c)} Eval on longer videos}
    \vspace{-1em}
    \caption{(a) We achieve comparable validation loss to the full-attention baseline and outperform both Mamba with sliding window and sliding window attention baselines. This improvement over SWA is consistent across different window sizes (b) and when evaluating on longer videos (c).}
    \label{fig:exp-video-baseline}
    \vspace{-0.2in}
\end{figure}

\noindent\textbf{Evaluation.}
We evaluate all models on a collection of 2,000 videos after 5,000 training iterations by computing the denoising loss at five timesteps (550, 650, 750, 850, 950). 
Figure~\ref{fig:exp-video-baseline} plots the chunk-wise denoising loss across evaluated video frames. We only measure validation loss  up to training sequence length. See project website for our autoregressively generated videos \footnote{See video results in project website: \url{https://tianyuanzhang.com/projects/ttt-done-right/}}. 

\subsection{Analysis on Design Choices}\label{sec:analysis}

\begin{figure}[htbp]
    \centering
    \includegraphics[width=\textwidth]{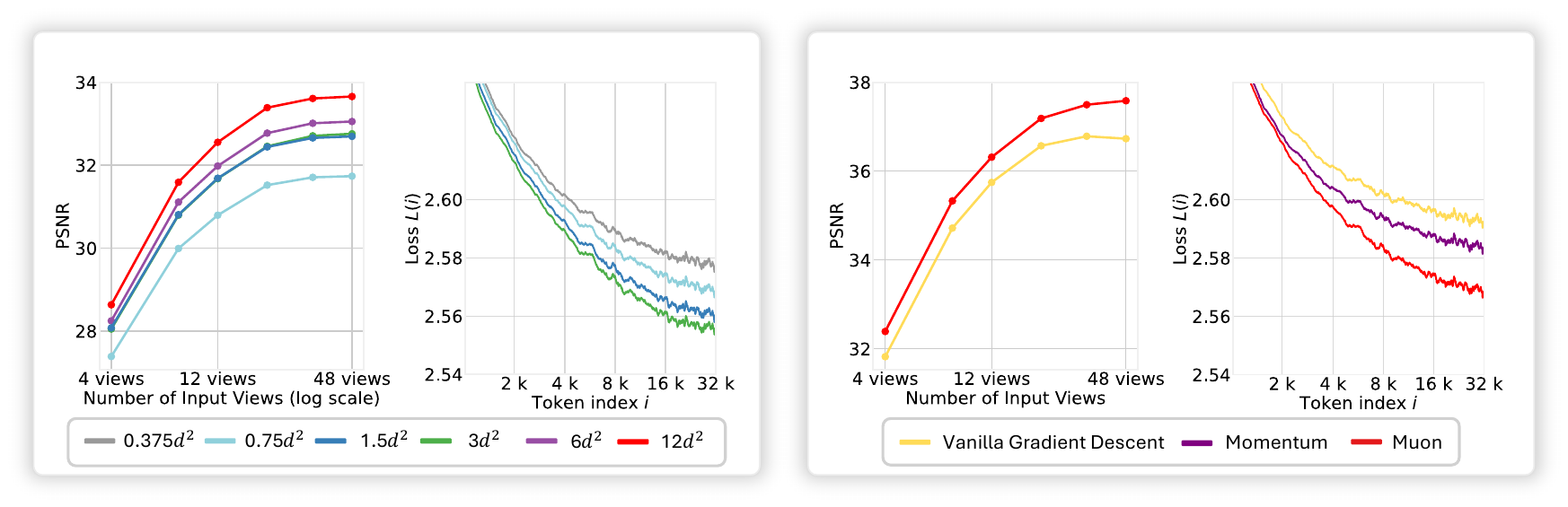}
    \put(-373, 113){\capfontscriptsize novel view synthesis}
    \put(-268, 113){\capfontscriptsize language model}
    \put(-173, 113){\capfontscriptsize novel view synthesis}
    \put(-70, 113){\capfontscriptsize language model}
    \put(-330, -5){\capfont \textbf{(a)} State Size Scaling}
    \put(-160, -5){\capfont \textbf{(b)} Different Test-Time optimizer}
    \caption{ \textbf{(a)} Scaling up the state size consistently improves performance in both novel view synthesis and language modeling tasks. Note, the largest version has state size of $12d^2$ per block, totaling 40\% of model weights as fast weights. \textbf{(b)} Comparison of test-time optimizers demonstrates Muon’s surprising effectiveness over Vanilla Gradient Descent and Momentum. }
    \label{fig:exp-state-optimizer}
\end{figure}
\begin{figure}[htbp]
    \centering
    \includegraphics[width=0.99\textwidth]{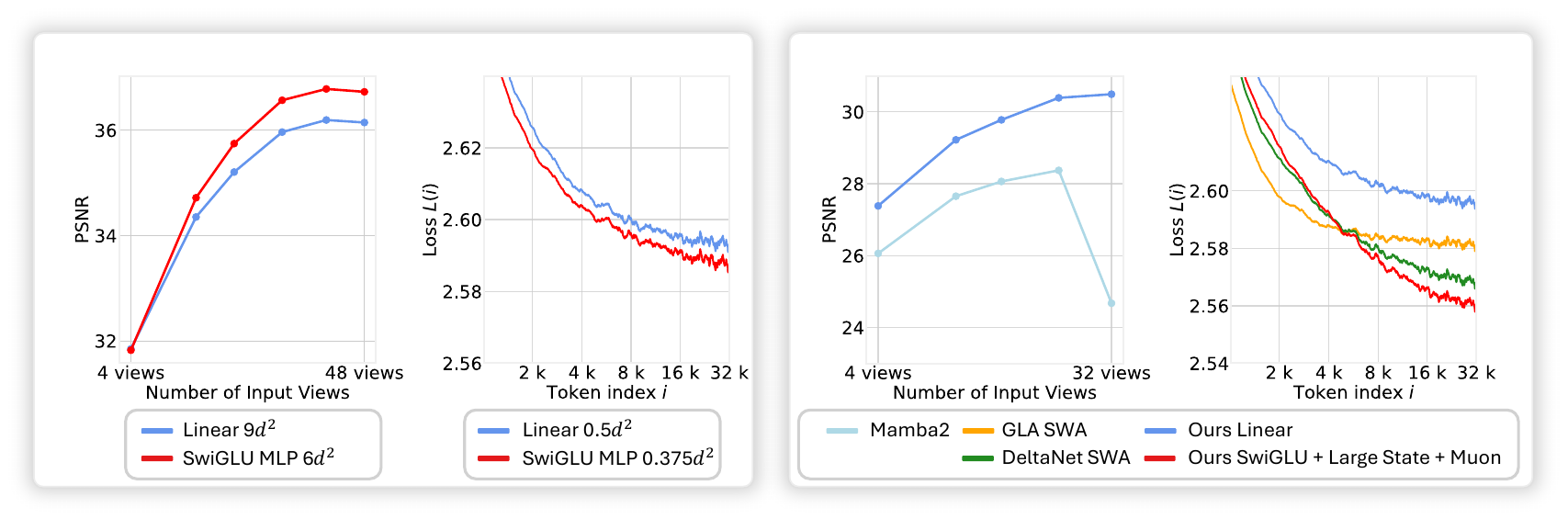}
    \put(-360, 115){\capfontscriptsize novel view synthesis}
    \put(-262, 115){\capfontscriptsize language model}
    \put(-173, 115){\capfontscriptsize novel view synthesis}
    \put(-75, 115){\capfontscriptsize language model}
    \put(-360,-5){\capfont \textbf{(a)} Linear v.s. NonLinear Fast weight}
    \put(-176,-5){\capfont \textbf{(b)} Large-Chunk v.s. Per-token Recurrence}
    \caption{\textbf{(a)} Nonlinear fast weights consistently outperform linear fast weights despite using smaller state sizes. \textbf{(b)} Our linear large-chunk recurrence approach significantly outperforms linear per-token recurrence (bidirectional Mamba2) for view synthesis tasks at the same state sizes.  In language tasks, linear large-chunk recurrence of the same state size underperforms the GLA baseline, but when combined with larger nonlinear states and Muon test-time optimizer, it surpasses all per-token recurrence methods.}
    \label{fig:exp-linear-token-recurrence}
\end{figure}
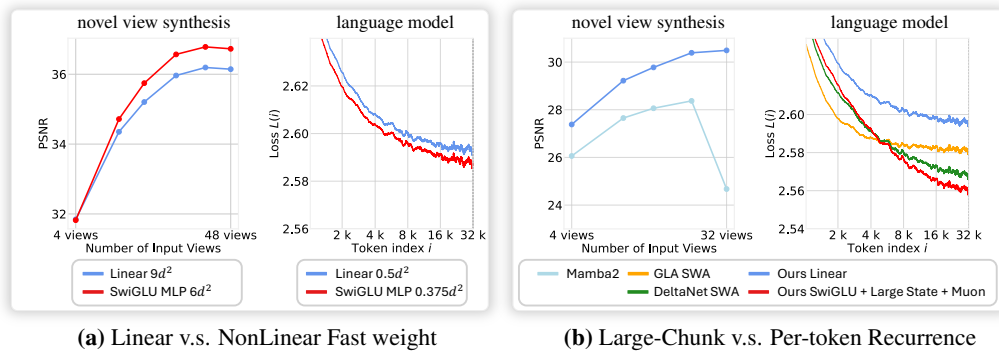

In this section, we analyze several key design choices in our model, focusing on both the novel view synthesis and language modeling tasks, where good metrics exist. Specifically, we evaluate the 
impact of state size (Fig.~\ref{fig:exp-state-optimizer}a), 
test-time optimizers (Fig.~\ref{fig:exp-state-optimizer}b),
linear versus nonlinear fast weights (Fig.~\ref{fig:exp-linear-token-recurrence}a), and 
per-token recurrence versus chunk-wise recurrence (Fig.~\ref{fig:exp-linear-token-recurrence}b).
Overall, we find that a large state size, advanced optimizers such as Muon, and nonlinear fast weights significantly improve our model’s performance. For comparing chunk recurrence with per-token recurrence, in a controlled NVS experiment, our linear large-chunk recurrence strategy outperforms linear per-token recurrence with the same state size. For language modeling, where chunk structures are not inherent, our linear large-chunk recurrence variant—while initially underperforming per-token methods like GLA and DeltaNet—surpasses them when combined with a large nonlinear state and the Muon optimizer.
We refer the readers to each figure and its caption for more detailed analysis.

\paragraph{State size scaling.} 
These controlled experiments utilize a SwiGLU MLP for fast weights and the Muon as the test-time optimizer.
For NVS, experiments were conducted on the object dataset. All models were trained for 167B tokens, using 14 stacked blocks and a model dimension $d=768$. To change the state size, we keep the head dimension fixed as model dimension. i.e. single head, and vary the intermediate dimension of SwiGLU MLP, such that the intermediate dimension ranges from $192$ to $3072$. The largest configuration results in a state size per model block as $12d^2$, totaling $40\%$ of model weights as fast weights.  For the language model experiments, we use the 760 milion parameter setup, where the chunk size and sliding window attention (SWA) window size were set to 2048 tokens. We keep the intermediate dimension of the fast weight SwiGLU MLP the same as the head dimension. We increase the state size while proportionally decreasing the number of heads to maintain a fixed model dimension. Figure~\ref{fig:exp-state-optimizer}(a) demonstrates that larger state sizes consistently improve performance. Notably, the performance gap between small and large state sizes widens with increasing sequence length.

\paragraph{Test-Time optimizer comparison. }
We compare Muon with vanilla Gradient Descent (GD) and GD with momentum. Details on momentum implementation are in Appendix~\ref{paragraph:momentum}.   For NVS, we train all compared approaches for 671 tokens using model specs of 24 stacked blocks with model dimension of 768. Language modeling experiments used the 760M parameter setup. Figure~\ref{fig:exp-state-optimizer}(b) shows Muon consistently outperforming other optimizers.

\paragraph{Linear v.s. NonLinear fast weight.} Our default fast weight function is a SwiGLU MLP without bias terms (nonlinear). We compare this against a simple linear fast weight, $f_W(x) = Wx$. Both are updated using the same online dot product loss for key-value association. Figure~\ref{fig:exp-linear-token-recurrence} (a) presents this comparison for NVS and language modeling. Although the linear fast weights were configured with a larger state size than the nonlinear SwiGLU, they achieved lower performance.  NVS models were trained for 671B tokens with 24 blocks and $d=768$.  Language modeling used the 760M parameter setup.

\paragraph{Large-chunk v.s. Per-token recurrence.} 
Figure~\ref{fig:exp-linear-token-recurrence}(b) presents controlled experiments comparing our large-chunk recurrence with per-token recurrence. 
In the novel view synthesis (NVS) task, ``Our Linear" variant employs a linear fast weight: $f_W(x) = Wx$ and is benchmarked against a Mamba-2 baseline (a linear per-token recurrence model) with an identical state size. To accommodate the bidirectional context required by NVS over input image tokens, the Mamba-2 baseline uses two Mamba-2 layers applied in opposite directions within each model block. Both our linear variant and this bidirectional Mamba-2 have state size of $d^2$ per block. Both of these two approaches employs a per-image window attention within each model block. Under this fair comparison, our linear large-chunk recurrence achieves significantly better view synthesis performance.

For the language modeling experiments also shown in Figure~\ref{fig:exp-linear-token-recurrence}(b), the blue line ``Our Linear" variant uses the same state size ($0.25 d^2$) as the GLA SWA baseline. It initially underperforms GLA SWA (blue line underperforms yellow line),  likely because language data lacks the inherent chunk structures that benefit our basic linear chunk recurrence. However, when LaCT is equipped with a larger non-linear state ($1.5d^2$) and Muon updates, we significantly outperform these per-token recurrence baselines.

\section{Related Work}
\label{appendix_sec:realted_work}



\paragraph{Test-time training.} Test-Time Training (TTT)~\cite{sun2024learning}  is an emerging concept in sequence modeling that extends the concept of recurrent states in RNNs to online-adapted neural network components. In TTT models, a subset of weights, termed "fast weights," are updated to learn in-context. Existing methods typically employ a self-supervised loss that encourages these fast weights to memorize key-value associations from in-context tokens, using variants of gradient descent for online adaptation. TTT~\cite{sun2024learning,wang2025testtimeregressionunifyingframework} has opened a vast design space for new recurrent model architectures. For instance, many recent works have developed novel test-time optimizers~\cite{behrouz2024titans, karami2025lattice} and online training objectives~\cite{behrouz2025s}. However, current TTT approaches often suffer from low hardware utilization and limited state sizes, and consequently have not yet demonstrated their full potential.  Our work primarily addresses these challenges by advocating for a new paradigm of using extremely large online minibatch (chunk) sizes for updating the fast weights. This paradigm can achieve orders-of-magnitude higher hardware utilization without relying on error-prone custom kernel implementations. Furthermore, it enables efficient scaling of nonlinear state sizes and offers the flexibility to use diverse fast weight neural networks and optimizers, thereby accelerating research progress in this area.

\paragraph{Combining chunk attention with recurrence.}
Several recent models combine local chunk attention with linear recurrence, such as Gated Attention Unit (GAU) \cite{hua2022transformerqualitylineartime}, MEGA \cite{ma2023mega}, MEGALODON \cite{ma2024megalodon}, and InfiniAttention \cite{munkhdalai2024leavecontextbehindefficient}. Among these, InfiniAttention is conceptually closest to our work, as it incorporates recurrence at the chunk level using the delta rule—interpreted as an online linear regression objective from the perspective of Test-Time Training (TTT). However, this update rule is limited in expressivity. In contrast, we employ a significantly more expressive update mechanism derived from a more general TTT framework, and demonstrate the substantial gains this brings.

Block-Recurrent Transformer \cite{hutchins2022blockrecurrent} also explores large chunk memory updates, where memory tokens act as recurrent states that can self-attend and cross-attend with input tokens during each chunk update via attention mechanisms.  The Perceiver-style register-token attention baseline used in our novel view synthesis experiments (Sec.~\ref{sec:exp_nvs}, Table~\ref{tab:attention_comparison}) is conceptually similar to the Block-Recurrent Transformer in its use of register tokens for context compression. As shown in Figure~\ref{fig:view_syn_results}, our method significantly outperforms this approach in both speed and quality, with a comparable state size.


\paragraph{Novel view synthesis.} Novel view synthesis (NVS) is a long-standing task at the intersection of computer vision, graphics, and computational photography, requiring algorithms to render images of a static scene from previously unobserved viewpoints. Optimization-based approaches, such as NeRF~\cite{mildenhall2021nerf} and 3D Gaussian Splatting~\cite{kerbl20233d}, have achieved significant breakthroughs.
These methods optimize a set of parameterized graphics primitives (i.e., explicit or implicit representations of radiance fields) through differentiable volumetric rendering to minimize reconstruction loss on input images. After an optimization process typically lasting tens of minutes, these approaches can render novel views photorealistically, and the optimized parameters form a 3D representation of the input scene.

Recently, data-driven approaches~\cite{zhang2024gs, jin2024lvsm, ziwen2024long, liu2023zero} have also shown promising results. These methods can either directly render novel views or predict 3D representations given input images. Although successful on simpler object datasets, these methods often struggle with densely sampled scenes (e.g., scenes with over 100 input images).  Our experiments demonstrate that our large-chunk test-time training approach outperforms or achieves comparable performance to 3D Gaussian Splatting on challenging scene datasets with up to 128 input images with $960\times536$ resolution at challenging scene datasets.We hope our method will inspire further research into effectively scaling data-driven NVS methods to longer and more complex input sequences.

\paragraph{Autoregressive video diffusion.}
Current state-of-the-art video generation is dominated by bidirectional diffusion transformers operating in latent space~\cite{videoworldsimulators2024, yang2024cogvideox, polyak2410movie, wang2025wan}. These methods factorize the video distribution into a sequence of conditional distributions based on noise levels, following diffusion processes~\cite{sohl2015deep, song2020score} or flow matching~\cite{lipman2022flow}, then use a diffusion transformer to jointly learn all the conditional distribution. Autoregressive video diffusion~\cite{alonso2024diffusion, jin2024pyramidal, valevski2024diffusion, ruhe2024rolling, yin2024slow, song2025history} introduces an additional temporal dimension to this factorization, where the neural networks learns to model the conditional probability of the next chunks of videos at different noise levels, conditional on previous videos and noisier  version of current video frames.  

During training, some autoregressive methods employ teacher forcing, supervising the model on noisy video frames given previous  clean context frames as condition~\cite{alonso2024diffusion,jin2024pyramidal,valevski2024diffusion}, though this can lead to low token utilization, i.e. only a small portion of tokens get supervision. To improve token efficiency, other techniques such as progressive noise injection~\cite{ruhe2024rolling} or the use of frame-independent noises (sometimes in a diffusion-forcing style)~\cite{yin2024slow, chen2024diffusion, magi1} have been proposed. When applying our large-chunk design to autoregressive video generation, we format the input sequence with interleaved clean and noisy chunks (see Equation~\ref{eq:video_sequence_basic}). This strategy achieves over 50\% token utilization and integrates effectively with our large-chunk TTT implementation, by only changing a few lines to constrain fast-weights are only updated on clean frame chunks.

\section{Limitation}
One limitation of our method is the absence of rotation invariance. Unlike softmax attention and linear attention, which remain invariant under uniform rotations of queries and keys (a property leveraged by relative positional encodings such as RoPE~\cite{su2023roformer}), our SwiGLU and Linear Fast Weight components do not exhibit this property. The practical implications of this absence remain underexplored.

We conduct our experiments on three tasks.
Although the tasks are diverse and cover different modalities, the effectiveness of our method would request of more tasks.
For example, the novel-view synthesis task is essentially a 3D reconstruction with input pose information.
The task of unposed reconstruction is more challenging and is not explored in this paper.

On the language modeling task, some key aspects are not explored due to computation limitation.
These aspects include the reasoning capacity of our \methodname{} model and also the scalability regarding the parameter size.
Previous papers showed that a main weakness of the state-based model (where \methodname{} belongs to) is its reasoning ability.
However, the reasoning ability is only gained with certain amount of training compute thus it is beyond our budget.

Lastly, for the autoregressive video diffusion, it is hard to find a reliable and distinguishable metric to measure the model's scalability.
It is in contrast to the language modeling with perplexity (i.e., log likelihood loss) and the novel-view synthesis with PSNR.
We show the validation loss in our paper and it is a common choice in evaluating the scalability of video generation.
This is a general problem for the video generation evaluation and is not specific to our paper.

\section{Conclusion} \label{sec:conclusion}
We presented \methodname{}, a novel model architecture that integrates large-chunk test-time training for capturing long context with window attention for modeling local structure. 
We validated \methodname{} across three diverse tasks spanning different modalities---novel view synthesis, language modeling, and autoregressive video diffusion---and demonstrate its effectiveness  
by achieving superior or competitive performance when compared to state-of-the-art baselines.
\methodname{} achieves high GPU efficiency even with native PyTorch implementation with dozens of lines of code and supports efficient scaling
up of the state size and more flexible designs in test-time training models and optimizers.  
By open-sourcing the code and weights, we hope that \methodname{} can advocate future research explorations 
into more performant architectures for long-context modeling.

\section*{Acknowledgment}

We thank Ziwen Chen for processing the DL3DV dataset and providing the K-means clustering.
We thank Nathan Carr, Feng Liu, Jianming Zhang, and Hailin Jin for their generous support on this project. 
We thank Haian Jin and Zexiang Xu for leading the LVSM project. 
We thank Baqiao Liu and Haoran Cai for the video data loader. We thank Guo Han for details on the language dataset. We thank Jeremy Bernstein for discussions on Muon. 

{\small
\bibliographystyle{unsrt}
\bibliography{main}
}

\newpage
\appendix



\section{\methodname{} Model Implementation Details }
\label{sec:appen_lact_details}

\begin{algorithm}[p]
\caption{Large Chunk Test-Time Training Layer Pseudocode}\label{alg:full_lact_layer}
\newcommand{\hlbox}[1]{%
  \fboxsep=1.2pt\hspace*{-\fboxsep}\colorbox{blue!10}{\detokenize{#1}}%
}
\lstset{style=mocov3}
\begin{lstlisting}[
    language=python,
    escapechar=@,
    label=code:lact_full]
    def apply_fw(fast_weight, q):
        w1, w2, w3 = fast_weight
        hidden = silu(matmul(q, w1)) * matmul(q, w3) # [b, l, dh] = [b, l, d] x [b, d, dh]
        return matmul(hidden, w2)
        
    def update(fast_weight, k, v, lr, use_muon=True):
        """
        Fast-weight update for a SwiGLU MLP using chunk of tensors. 
        Args:
            fast_weight : tuple(w1, w2, w3) with shapes: w1, w3: [b, d, dh]; w2: [b, dh, d]
            k, v        : key / value tensor of shape [b, l, d]
            lr:         : per-token learaning rates of shape [b, l, 3] -> (lr1, lr2, lr3)
            use_muon    : weather to apply Muon to orthogonalize the update
        Note:
            The head dimension for input tensors k, v, lr are assumed to be merged into the batch dimension. This is not necessary, but simplifies shape annotation in this pseudocode.
        """
        # Forward with k:
        gate_before_act = matmul(k, w1) # [b, l, dh] = [b, l, d] x [b, d, dh]
        hidden_before_gate = matmul(k, w3) # [b, l, dh] = [b, l, d] x [b, d, dh]
        hidden = silu(gate_before_act) * hidden_before_gate
        
        # Backward:
        dhidden = matmul(v, w2.transpose(-1, -2)) # [b, l, dh] = [b, l, d] x [b, d, dh]
        dhidden_before_gate = dhidden * silu(gate_before_act)
        dgate = dhidden * hidden_before_gate
        dgate_before_act = silu_backprop(dgate, gate_before_act)

        # Compute gradients: 
        w2.grad = -matmul(hidden.transpose(-1, -2), v * lr2) # [b, dh, d] = [b, dh, l] x [b, l, d]
        # [b, d, dh] = [b, d, l] x [b, l, dh]
        w1.grad = -matmul((k * lr1).transpose(-1, -2), dgate_before_act)
        w3.grad = -matmul((k * lr3).transpose(-1, -2), dhidden_before_gate)

        # Weight update
        if use_muon:
            for w in fast_weight:
                w.grad = zeropower_via_newtonschulz5(w.grad)
        for w in fast_weight:
            w = (w - w.grad) / (w - w.grad).norm(dim=1) * w.norm(dim=1)
        return fast_weight

    def silu_backprop(dy, x):
        sigma = sigmoid(x)
        return dy * sigma * (1 + x * (1 - sigma))

    #################################### MultiHead LaCT layer ####################################
    # x: input sequence [b, l, d], b is the batch dim, l is length, d is model dimension
    # fast_weight: tuple of initial fast weights-(w1, w2, w3); w1, w3 of shape [nh, d, dh], w2: [nh, dh, d]
    # ttt_config: list of (operation, begin, end) tuples
    qkv = silu(LinearQKV(x)) # [b, l, d * 3]
    qkv = rearrange(qkv, `b l (nh hd) -> (b nh) l hd`, nh=num_heads).split(3) # merge heads into batch dim
    q, k = q / q.norm(-1), k / k.norm(-1)
    lr = softplus(LinearLR(x) + const_lr_bias) # [b, l, 3 * num_heads]
    lr = rearrange(lr, `b l (nh 3) -> (b nh) l 3`, nh=num_heads)
    fast_weight = repeat(fast_weight, dim=0, repeat=b) # [nh, ...] -> [b * nh, ...]
    
    o = zeros_like(v) # [b * nh, l, hd]
    for mode, begin, end in ttt_config:
      qi, ki, vi, lri = q[:, begin:end], k[:, begin:end], v[:, begin:end],  lr[:, begin:end]
      
      if mode == "update_then_apply": # figure-3(a, b) bidirectional attention
        fast_weight = update(fast_weight, ki, vi, lri, use_muon)
        o[:, begin: end] = apply_fw(fast_weight, qi)
        
      elif mode == "apply_then_update": # figure-3(b) shifted block-wise causal
        o[:, begin: end] = apply_fw(fast_weight, qi)
        fast_weight = update(fast_weight, ki, vi, lri, use_muon)

      elif mode == "update_only":
        fast_weight = update(fast_weight, ki, vi, lri, use_muon)
      
      elif mode == "apply_only":
        o[:, begin: end] = apply_fw(fast_weight, qi)

    o = RMSNorm(o) # per-head norm
    o = LinearOutput(rearrange(o, `(b nh) l hd -> b l (nh hd)`, nh=num_heads))
    return o
\end{lstlisting}
\end{algorithm}

\begin{algorithm}[t]

\caption{LaCT Layer with In-Layer Hybrid Window Attention Pseudocode}\label{alg:lact_hybrid}
\newcommand{\hlbox}[1]{%
  \fboxsep=1.2pt\hspace*{-\fboxsep}\colorbox{blue!10}{\detokenize{#1}}%
}
\lstset{style=mocov3}
\begin{lstlisting}[
    language=python,
    escapechar=@,
    label=code:gslrm]
    # Input: 
    # x: input sequence [b, l, d], b is the batch dim, l is length, d is model dimension
    # fast_weight: tuple of initial fast weights-(w1, w2, w3); w1, w3 of shape [d, dh], w2 of shape [dh, d]
    # ttt_config: list of (operation, begin, end) tuples

    q, k, v = LinearQKV(x).split(3)
    
    #### Local quadratic-cost window attention
    attn_q = q * learnable_q_scale + learnable_q_offset # per-channel rescale and shift
    attn_k = k * learnable_k_scale + learnable_k_offset # per-channel rescale and shift
    attn_o = local_softmax_multihead_attn(attn_q, attn_k, v, attn_mask)
    
    #### large chunk test-time training for long memory
    q, k = rearrange(q, k, `b l (nh hd) -> (b nh) l hd`, nh=num_heads)
    q, k = silu(q), silu(k)
    q, k = q / q.norm(-1), k / k.norm(-1)
    lr = softplus(LinearLR(x)) # [b, l, 3 * num_heads]
    lr = rearrange(lr, `b l (nh 3) -> (b nh) l 3`, nh=num_heads)

    # Perform update and apply_fw operations iteratively over chunks of tokens.
    lact_o = lact(fast_weight, q, k, v, lr, ttt_config)
    lact_o = RMSNorm(lact_o)
    
    scale_per_head = rearrange(silu(Linear(x)), `b l nh -> (b nh) l 1`, nh=num_heads)
    lact_o = lact_o * scale_per_head
    lact_o = rearrange(lact_o, `(b nh) l hd -> b l (nh hd)`, nh=num_heads)

    #### Merge attention results (shape: [b, l, d])
    o = attn_o + lact_o

    o = LinearOutput(o)

    return o
\end{lstlisting}
\end{algorithm}


\paragraph{State Size calculation.} Motivated by recent progress in LLM, we adopt SwiGLU-MLP~\cite{shazeer2020glu} without bias terms as the fast-weight 
network. 
Our fast weights consists of three weight matrix $W = \{W_1, W_2, W_3\}$, and the forward pass of the fast weight model is:
\begin{equation}
f_W(x) = W_2 \left[ \mathrm{SiLU}(W_1 x) \circ (W_3 x) \right]
\label{eq:swiglu_supp}
\end{equation}
where $\circ$ is an elementwise multiplication. We define $\mathit{hd}$ as the head dimension, $\mathit{nh}$ as the number of heads, and the intermediate dimension of the SwiGLU-MLP as $\mathit{hd} \times r$, where $r$ is a scaling multiplier.When $r > 1$, the intermediate dimension surpasses the input head dimension, which is the current common practice in LLMs.  TThus, matrices $W_1, W_2 \in \mathbb{R}^{\mathit{hd} \times \mathit{hd}}$ and $W_3 \in \mathbb{R}^{\mathit{hd} \times \mathit{hd} \times r}$. Consequently, the total state size becomes $\mathit{nh} \times \mathit{hd} \times \mathit{hd} *r$.  Given that typically the total head dimension across all heads equals the model dimension $d$ (i.e., $\mathit{nh} \times \mathit{hd} = d$), the total state size simplifies to:
\begin{equation}
    \text{State Size} = \frac{d^2}{\mathit{nh}} * r.
    \label{eq:state_size}
\end{equation}
Therefore, we can increase the state size either by reducing the number of heads or by increasing the intermediate dimension multiplier.

\paragraph{FLOPs calculation.} When using then negative dot product loss as the online test-time training objectives, we don't need to compute the final results of $f_W(v)$. We only need to compute $W_1v, W_3v$ when running forward pass with keys $k$, thus there are two matmuls in the forward pass with keys.  When computing the gradients, there are four matmuls. And in the final forward pass the queries, there would be three matmuls. So the total FLOPs with $n$ tokens would be:
\begin{equation}
    \text{FLOPs} = 4n \frac{d^2}{\mathit{nh}}r +  8n \frac{d^2}{\mathit{nh}}r +  6n \frac{d^2}{\mathit{nh}}r  = 18n \frac{d^2}{\mathit{nh}}r = 6 * \text{State Size} 
\end{equation}

\paragraph{Model initializations.} We randomly initialize the linear layers using a standard deviation of 0.02. For the learnable initial fast weights, we initialize them with a standard deviation of $1.0 / \sqrt{\text{fan-in}}$. Specifically, in the SwiGLU FFN fast weights, the matrices $w_1$ and $w_3$ have their fan-in set as the head dimension, while the fan-in of $w_2$ is the intermediate dimension of the SwiGLU FFN fast weights. Additionally, when local window attention is incorporated within the \methodname{} layer, we introduce four extra learnable embeddings: two scales and two reshifts for queries and values. We initialize the scale embeddings as ones and the reshift embeddings as zeros.

\paragraph{Details of Muon.} Muon~\cite{jordan2024muon} is a recently proposed optimizer that orthogonalizes the matrix gradients during updates of matrix weights. It utilizes Newton-Schulz iterations to achieve orthogonalization. Given a matrix gradient $G$, Muon first normalizes it as $G_0 = G / |G|F$, then iteratively applies:
\begin{equation}
\mathbf{G}_k = a\mathbf{G}_{k-1} + b(\mathbf{G}_{k-1}\mathbf{G}_{k-1}^{\mathrm{T}})\mathbf{G}_{k-1} + c(\mathbf{G}_{k-1}\mathbf{G}_{k-1}^{\mathrm{T}})^2\mathbf{G}_{k-1},
\label{eq:muon_iteration}
\end{equation}
where the constants $a, b, c$ are carefully chosen for optimal convergence. Following the original implementation, we set $a=3.4445$, $b=-4.7750$, $c=2.0315$, and perform five iterations.

Each Muon iteration requires three matrix multiplications, resulting in a computational cost per fast weight head of $2 \mathit{hd}^3 r + 2 \mathit{hd}^3 + 2 \mathit{hd}^3 r = \mathit{hd}^3 (4r + 2)$ FLOPS. Hence, the total computation for five iterations across all fast weights is:
\begin{equation}
5 \times \mathit{nh} \times \mathit{hd}^3 \times (4r + 2).
\end{equation}

For the case where $r=1$ (head and intermediate dimensions are equal), the total computational cost simplifies to:
\begin{equation}
30 \times \mathit{nh} \times \mathit{hd}^3 = 30 \times \mathit{hd} \times \text{State Size}.
\end{equation}
This indicates that the computational overhead of Muon becomes less significant than computing token outputs only if the online chunk size exceeds $\frac{5}{3} \mathit{hd}$.

\paragraph{Rotation invariance.} Softmax attention and linear attention exhibit rotation invariance: rotating the queries and keys by the same rotation matrix does not alter the output. This property is also used in developing relative positional encodings, like RoPE~\cite{su2023roformer}. In contrast, our SwiGLU and Linear Fast Weight components do not possess this property.

\paragraph{Implementing momentum for test-time optimizers.}  \label{paragraph:momentum}
Muon uses momentum by default. Following Titans~\cite{behrouz2024titans}, we implement momentum in the test-time optimizer by predicting a scalar momentum $\beta_i$ from each token: 
\begin{equation}
    \beta_i = \sigma(\text{Linear}(\boldsymbol{x_i})), 
\end{equation}
where $\sigma$ is the sigmoid function.  This $\beta_i$ is then averaged over all tokens in the chunk, and the average momentum is applied as follows:
\begin{equation}
\begin{aligned}
    g \leftarrow&  \sum_i^b \eta_i \nabla_W \mathcal{L}(f_W(k_i), v_i), \\
    M \leftarrow&  M (\sum_i^b \beta_i / b) + g, \\
    W \leftarrow& \text{weight-update}(W, M),
\end{aligned}
\end{equation}
where the $\text{weight-update}$ can be simple subtraction followed by L2 normalization normalization (as in Equation~\ref{eq:fast_weight_normalization_nomuon}). or Muon update before subtraction (as in Equation~\ref{eq:fast_weight_normalization}).

\subsection{Pseudocode}
\label{sec:ttt_pseudocode}

See Algorithm~\ref{alg:full_lact_layer} for pseudocode of a full \methodname{} layer. 
For details on how to mix local window attention inside each layer with shared query, value, embedding, see Algorithm~\ref{alg:lact_hybrid} for pseudocode.

\section{\methodname{} Architecture Details for N-Dimensional Data }
\label{sec:appen_model_arch}

\subsection{\methodname{} Architecture for Novel View Synthesis}
\label{sec:appen_nvs_model_arch}

Novel view synthesis (NVS) renders images of a static scene from novel viewpoints. 
Formally, given a set of $N$ input posed images $\{ (  \img_i, P_i ) \}_{i=1}^N$ of a static scene, 
where $I_i \in \mathbb{R}^{H\times W \times 3}$ is an RGB image and $P_i$ is its corresponding camera pose,
the model needs to synthesize new images from novel camera poses $P_\text{novel}$ that typically do not overlap with the input views.

Traditional methods in 3D vision usually solved the NVS task by reconstruction and rendering. 
The reconstruction compresses the posed input into a compact representation and then the render renders the novel view from it.
Our method mimics such pipeline, where we first compress the posed input images into fast weights by the `Update Operation' (Sec~\ref{sec:ttt_layer}) in \methodname{}.
Then we render the novel view images from the information in the compressed weights with the `Applying Operation'.

In details, we first convert the input and output into tokens.
The camera pose $P$ for each view is represented in dense ray information for each pixel (usually from the camera intrinsics and extrinsic), i.e., $P = (\text{rays}_o, \text{rays}_d)$.
$\text{rays}_o \in \mathbb{R}^{H\times W \times 3}$ is the 3D coordinate for the origins of the ray, and $\text{rays}_d \in \mathbb{R}^{H\times W \times 3}$ is the direction of the ray.
We follow GS-LRM~\cite{zhang2024gs} to use the Pl\"ucker ray embedding for the rays.
Pl\"ucker ray embedding computes the cross product between the ray origin and ray direction for a normalization.
The final positional embedding is a concatenation of the ray's origin, the ray's direction, and the cross product of the above two:
$[\text{rays}_o, \text{rays}_d, \text{rays}_o \times \text{rays}_d]$.
We add ray's origin into the embedding since different origins in the same ray can results in different colors due to the occlusions.
We then use patchifying and two different Linear layers to convert the RGB map and ray map (i.e., the positional embedding) into tokens.
For the posed RGB input images, we simply the sum the RGB embedding and pose embedding as model input.
For the novel view cameras, we only use the pose embedding.

\begin{figure}[t!]
    \centering
    \includegraphics[width=0.8\textwidth]{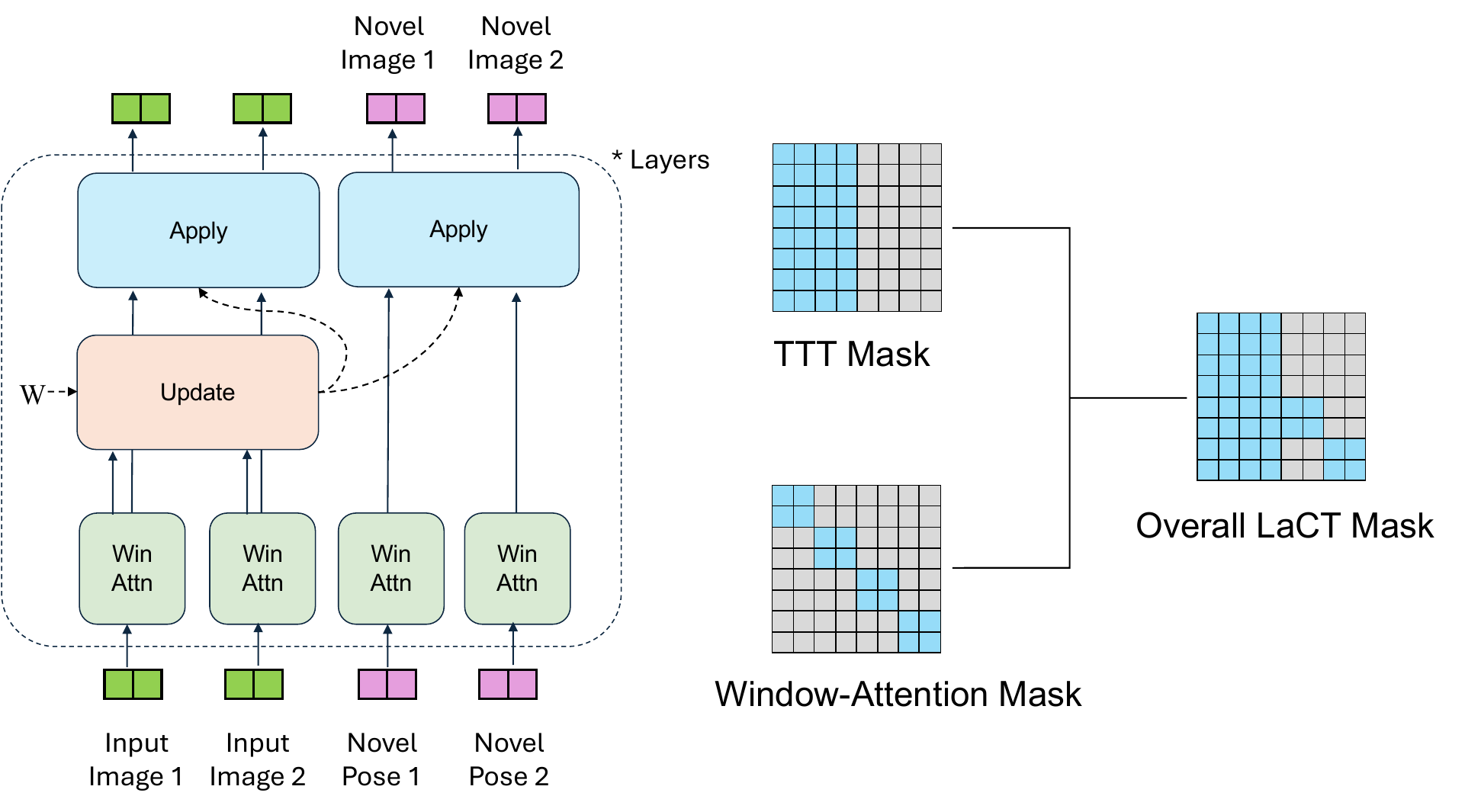}
    \caption{
    Detailed \methodname{} model for our Novel-View Synthesis. Dashed line indicates flow of fast weight. Solid line indicates flow of tokens.
    Window attention is bidirectional within a single image, either the input image or the novel target image.
    TTT updates over all input tokens and apply to all tokens.
    }
    \label{fig:appen_lvsm_model}
\end{figure}

We illustrate the design of NVS's \methodname{} block in Fig.~\ref{fig:appen_lvsm_model}. 
We first apply the attention for each image (either input or the novel target). 
The attention is bidirectional for the tokens belonging to the same image, and is independent among different images. 
Then, the TTT update operation is applied to all input tokens, i.e., all tokens that belonging to all input images.
The updated weight then is applied to all tokens. 
The two updates blocks in Fig.~\ref{fig:appen_lvsm_model} take the same updated fast weight thus can be combined, and we left two `Apply' block for clearance.
Note that the original NVS task definition renders novel views independently.
We here supervise multiple novel image poses and their corresponding images in a single data point for better training efficiency.
Given the design, the novel images are independent to each other, which is illustrated in the `Overall LaCT Mask' in the right of Fig.~\ref{fig:appen_lvsm_model}.
The layer normalization layer, the residual connections, and the feed-forward network is omitted for clarity.
The block is repeated by number of layers times to formulate the full model.
The general model largely follow the design of the encoder-decoder model in LVSM~\cite{jin2024lvsm}, except we use TTT in replace of transformer for long-context modeling.

For actually using this model for NVS task during inference, we first get the updated fast weight with all input images.
Then, we would not change the fast weight during the rendering process (i.e., the process to convert novel camera poses to the novel images).
The \methodname{} during rendering would be similar to a ViT (Vision transformer) architecture despite having two feed-forward networks:
the feed-forward network from the fast weight stores the scene information, and the feed-forward network from the slow weight (i.e., the FFN in Fig.~\ref{fig:lact_model_arch}) stored the world knowledge like physical rendering rules.

\begin{figure}[t!]
    \centering
    \includegraphics[width=1.0\textwidth]{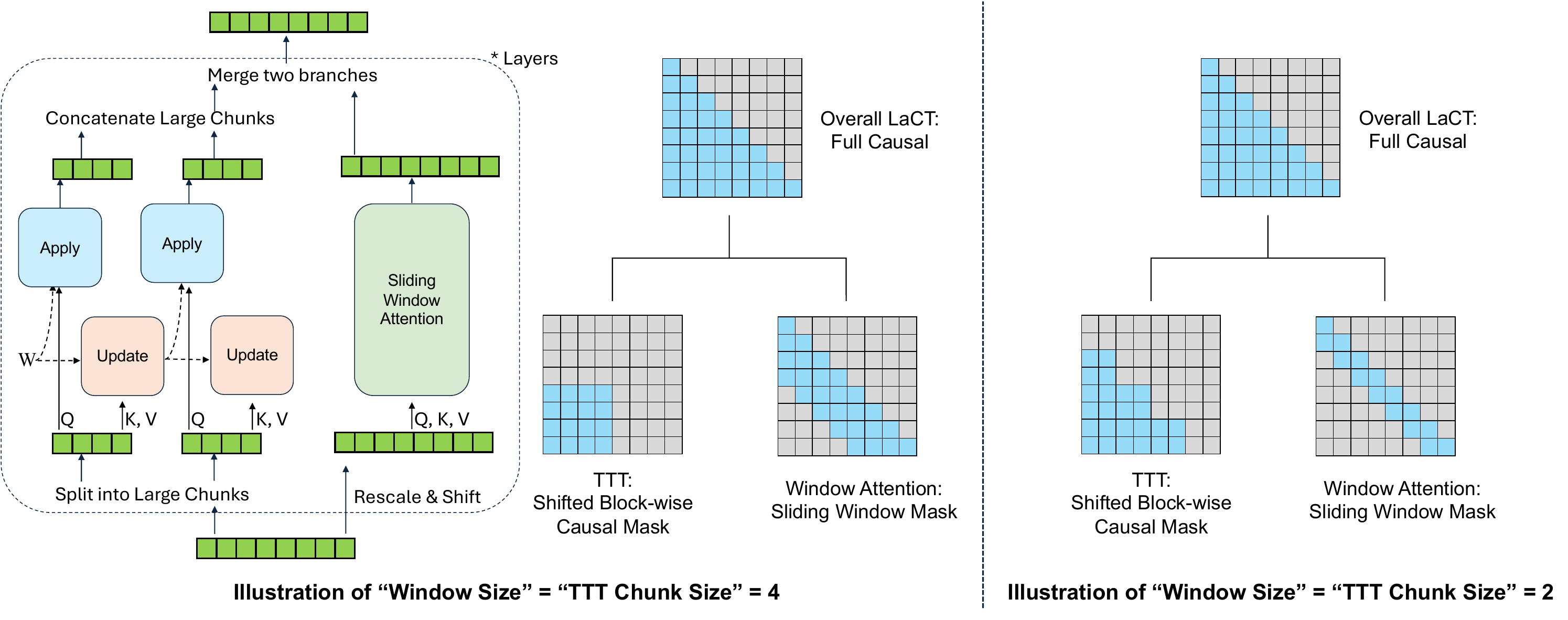}
    \vspace{-0.25in}
    \caption{
    Detailed \methodname{} model for language models. Dashed line indicates flow of fast weight. Solid line indicates flow of tokens. 
    We illustrate with TTT chunk size 4 or 2, and the actual chunk size is over 2048 in \methodname{}.
    We take the parallel design for the window attention and TTT block with shared QKV.  
    The overall mask is the causal mask.
    }
    \vspace{-0.15in}
    \label{fig:appen_lm_model}
\end{figure}

\subsection{\methodname{} Architecture for Language Models}
\label{sec:appen_lm_model_arch}
Autoregressive Language Models (LM)  predicts the distribution of the next tokens $p_{\theta}(x_n|x_1, \dots, x_{n-1})$  from its history context.
It is a factorization of the full sequence distribution $p_{\theta}(x_1, \dots, x_n)$ through chain rule $p_{\theta}(x_1, \dots, x_n) = p_{\theta}(x_1)  p_{\theta}(x_2 | x_1) \ldots p_{\theta}(x_n|x_1, \dots, x_{n-1})$.
Thus it requires a token-level causal mask (demonstrated in the topright of Fig.~\ref{fig:appen_lm_model}) 
and this is the main difficulty for the large-chunk design in \methodname{}.
We use a combination of TTT layer with `Shifted Causal Block Mask' (introduced before in Fig.~\ref{fig:large_chunk_recurrence}c) and a sliding window attention to facilitate it.
By shifting the mask of TTT, it excludes the information leakage from future tokens.
As shown in the right part of Fig.~\ref{fig:appen_lm_model}, the overall dependency mask is the union of the TTT mask and the sliding window attention mask.
To achieve a token-level causal mask without bubbles, the only requirement is that the window size of the sliding window attention is greater or equal to the chunk size from the TTT.
We illustrate two example of such mask with `Window Size' = `TTT Chunk Size' = 2 or 4.
The actual chunk size and attention window size is above 2048 in our implementation for better utilization and state size scaling (discussed in Sec.~\ref{sec:efficiency_challenge}).

As illustrated in the left most of Fig.~\ref{fig:appen_lm_model}, we employ a parallel design of the TTT layer and sliding window attention to save the number of model parameters and computation FLOPs.
In details, the query (Q), key (K) and value (V) are shared between the TTT layer and window attention.
Sliding window attention is an attention with constant window size over the past history, starting from the target tokens.
For the TTT layer, we start with an apply operation over the first chunk using the initialized fast weight (i.e., unupdated yet).
The `apply' operation is followed by the `update' operation over the first chunk.
In this way, the `apply' operation would not see information inside the current chunk to avoid leaking future token information inside the chunk.
Alternatively using `apply' followed by `update' over subsequent blocks completes the desired `shifted block-wise causal mask' illustrated in Fig.~\ref{fig:appen_lm_model}.
For details of the parallel design, please refer to the Pseudocode Algorithm~\ref{alg:lact_hybrid}.


We use the multi-head design for both TTT layer and window attention, although their number of heads are different.
We empirically take less number of heads for TTT layer (i.e., larger head dimension) to enable larger state size, as state size is propotional to the head dimension in our design (Sec.~\ref{sec:appen_lact_details}, and Equation~\ref{eq:state_size}). By default, we use four heads in langugae model experiments.  For positional encoding, we use the same RoPE as the window attention branch.

\begin{figure}[t!]
    \centering
    \includegraphics[width=1.0\textwidth]{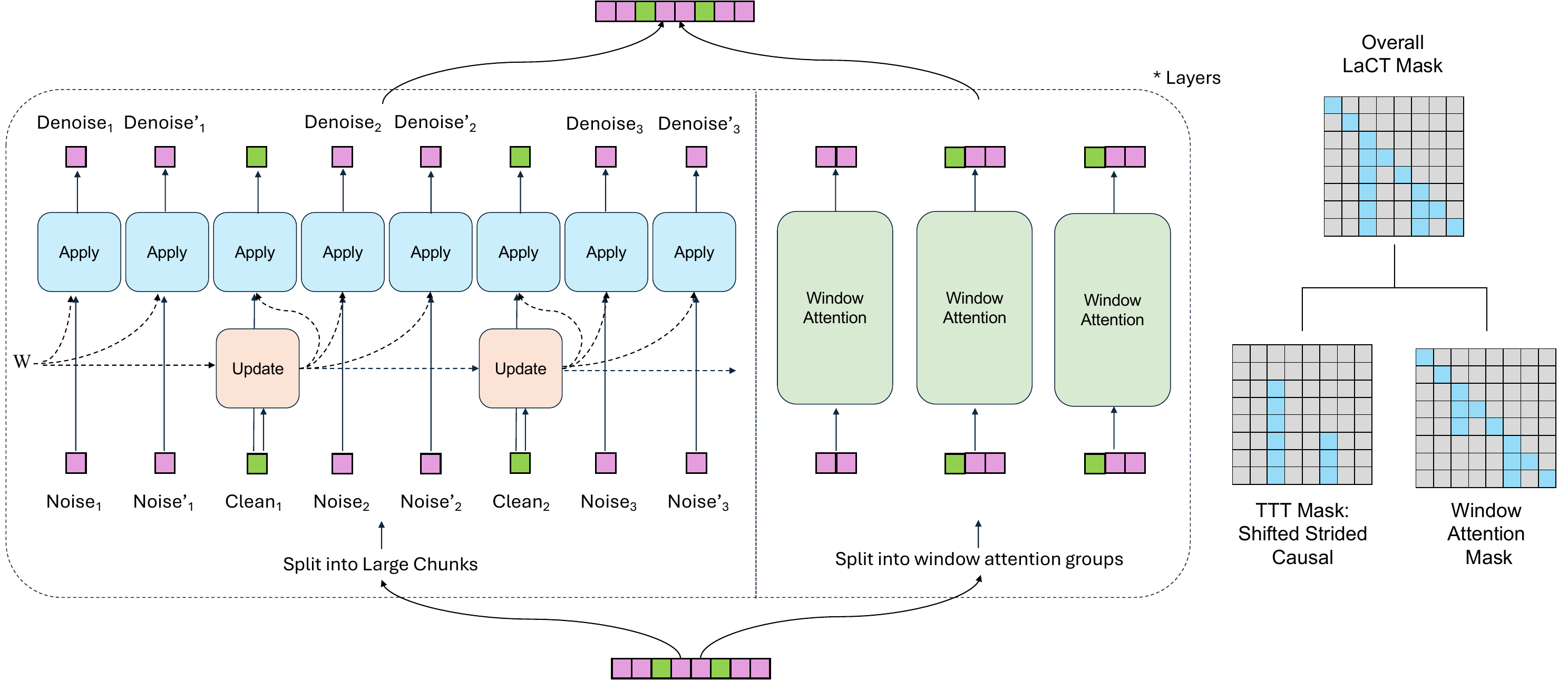}
    \vspace{-0.25in}
    \caption{
    Detailed \methodname{} model for autoregressive video generation by diffusion model. 
    The purple tokens are noisy frame chunk for training the diffusion model.
    The green tokens are clean frame chunk.
    Each token is a large chunk in TTT, e.g., 3 consecutive frames with 4680 tokens in total.
    Noise and Noise' are two noisy frames with independent Gaussian noise and time stamps over the same clean frame.
    We denoise them simultaneously to improve the utilization.
    Dashed line indicates flow of fast weight. Solid line indicates flow of tokens. 
    We take the parallel design for the window attention and TTT block with shared QKV.  
    The TTT mask is shifted  (i.e., started with apply) strided causal.
    The window attention excludes the attention from clean frame to future noisy frame, and also excludes the attention between the independent noisy frames.
    }
    \vspace{-0.15in}
    \label{fig:appen_video_model}
\end{figure}

\subsection{\methodname{} Architecture for Autoregressive Video Diffusion}
\label{sec:appen_video_model_arch}

Chunkwise autoregressive video diffusion generates videos by iteratively denoising sequential chunks of video frames, conditioned on previously generated clean frames. Each chunk can contain several video frames and span thousands of visual tokens.
We use teacher-forcing training by interleaving noisy and clean frame chunks. Specifically, a video of N frame chunks is structured as:
\begin{equation}
    S = [X_1^{\text{noise}}, X_1, X_2^{\text{noise}}, X_2, \ldots, X_N^{\text{noise}}]
    \label{eq:appen_video_sequence_basic}
\end{equation}
where each noisy chunk $X_i^{\text{noise}}$ is produced by adding unit Gaussian noise $\epsilon$ to the $i$-th clean video chunk as $X_i^{\text{noise}} = X_i (1 - t_i) + \epsilon t_i$ and $t_i \in [0, 1]$ denotes the strength of chunk-independent noise.  However, compared to previous methods that employs progressive \cite{ruhe2024rolling} or frame-independent noise strategies \cite{yin2024slow} like diffusion forcing \cite{chen2024diffusion},
our teacher-forcing formulation in Equation~\ref{eq:appen_video_sequence_basic} only uses around $50\%$ of the tokens of the entire sequence to compute the denoising loss. To improve token utilization, we consider an alternate approach by repeating each video chunk with two noise levels in the training sequence as:
\begin{equation}
    S = [X_1^{\text{noise}}, X_1^{\text{noise}_*},  X_1, X_2^{\text{noise}}, X_2^{\text{noise}_*},  X_2, \ldots, X_N^{\text{noise}}, X_N^{\text{noise}_*} ],
    \label{eq:video_sequence}
\end{equation}
where $X_i^{\text{noise}}$ and $X_i^{\text{noise}_*}$ represent two different noise levels applied to each clean video chunk $X_i$. This increases token utilization from $50\%$ to about $67\%$. While more repetition could further increase token utilization, it would also reduce training sample diversity; thus, we limit the repetition to twice. We use such repeating strategy when training the 1.3 billion parameter video diffusion model on five seconds videos. 

To process these interleaved noisy and clean chunks,  \methodname{} fast weights are updated exclusively using the clean video chunks. These updated weights are then applied to the current clean chunk and subsequent noisy chunks. The integrated local window attention uses a window size of two frame chunks and employs a block-wise causal mask. This mask allows noisy chunks to attend only to themselves and the immediately preceding clean chunk. By doing this, we main bidirectional dependies within each chunk and causal dependency across chunks. 

Similarly to our language model experiments, we integrate the TTT and local window attention into the same layer. Figure~\ref{fig:appen_video_model} illustrates this design for autoregressive video generation. The input sequence depicted in the figure follows Equation~\ref{eq:video_sequence} with each video chunk repeated twice with different noise levels. The pink color in the figure indicates noisy chunks and the green color indicates clean chunks.

\section{Experimental Details}
\label{sec:appen_experiment_details}

\subsection{Novel View Synthesis}
\label{sec:appen_nvs_details}

\noindent\textbf{Datasets \& metric.} We evaluate our approach on 
both object-level and scene-level datasets.
We use Objaverse dataset \cite{deitke2023objaverse} for object-level training, 
and render 32 random views per object,   
following the setup from LVSM \cite{jin2024lvsm} and GS-LRM \cite{zhang2024gs}.  
After training, we perform evaluations on the Google Scanned Objects (GSO) dataset \cite{downs2022google}, at resolutions of $256\times256$ and $512\times512$.
To ensure stablized evaluation results, we render at fixed view points instead of random view points as in training.
Each evaluation involves 4–48 input views and 8 novel views per object.
For scene-level evaluations, we adopt the challenging DL3DV scene dataset~\cite{ling2024dl3dv}, which contains over 11K training scenes and 140 testing scenes, each with approximately 300 views.
Evaluations are performed at a resolution of $960\times536$~\footnote{The original DL3DV 960p frames released in resolution of $960\times536$. To accommodate the patch-size $8$ in our modeling, we crop it to $960\times536$ and the camera parameters are changed accordingly.  }.
We report Peak Signal-to-Noise Ratio (PSNR) in the paper's main figures, and other metrics Structural Similarity Index Measure (SSIM) and LPIPS~\cite{zhang2018unreasonable} can be found in Tables below.
For DL3DV evaluation, we follow the original paper~\cite{ling2024dl3dv} and LongLRM~\cite{ziwen2024long} to use one frame from every $8$ frame in the full video sequence as the target frames.
The input frames are from the K-means clustering of all frames as in \cite{ziwen2024long}.

\noindent\textbf{Model details.}
Our models consist of 24 stacked \methodname{} blocks, each with a model dimension of 768. 
The detail of such block is illustrated in Sec.~\ref{sec:appen_nvs_details}:
Unless otherwise specified, we use a single-headed fast-weight SwiGLU-MLP with a hidden dimension of 1536. 
The window attention has 12 heads with head dimension 64, and is equipped with QK-normalization~\cite{henry2020query}.
The Feed-forward Network has 3072 as its intermediate hidden dimension.
The model has a total of 312M parameters, of which 84M are fast weights (i.e., $6 d^2$ per block).
We use an fast-weight lr initialization of $0.01$ by setting `const\_lr\_bias' in Algorithm~\ref{alg:full_lact_layer} to $\mathrm{softplus}(\text{const\_lr\_bias}) = 0.01$.
As we used Muon in fast weight update for NVS, \methodname{} is not sensitive to lr scale as discussed in Sec.~\ref{sec:update_rule}.

\noindent\textbf{Baselines.}
For object-level evaluations, we compare against two baselines, including a full-attention model, 
and a register-attention model in a Perceiver style~\cite{jaegle2021perceiver}, 
In the full-attention baseline, we 
replace the TTT layer in our model with a block-wise causal attention layer, where 
the input tokens interact bidirectionally and the novel view tokens cross-attend 
to the input tokens. Such a design resembles our method's prefill and parallel 
decoding strategy described in Section~\ref{sec:method-nvs}, and the key-value caches 
of the input tokens server as scene representations for novel view renderings. 
In the Perceiver-style model, we replace half of the TTT layers with 
input-to-register full-attention layers and the remaining half with register-to-novel-view
cross-attention layers. Such a model first compresses the input tokens into a constant 
set of register tokens and then decoding the novel view tokens by attending to the registers.
For scene-level evaluations, we compare against a state-of-the-art long-sequence 3D 
reconstruction work LongLRM~\cite{ziwen2024long} that applies Mamba~\cite{gu2023mamba} hybrid with full attention to predict 
3D Gaussian splats~\cite{kerbl20233d}. 
We also include comparisons with pure optimization-based 3D Gaussian splatting methods.
Tab.~\ref{tab:attention_comparison} compares the computational complexity of the 
baseline models and our models. 

\noindent\textbf{Training details.}
For object-level experiments, we first train all the model with $671$B tokens at 8 input view and 8 novel view setting at a resolution of $256\times256$. 
We then finetune them with $512\times512$ resolution for an additional $587$B tokens. 
For scene dataset, we first pre-train our model first with 32 input views and 32 novel views at $128\times128$ resolution for $1.5$T tokens, then progressively finetune at larger resolutions, larger field-of-views, and more input views.
The finetuning is always go with a non-squared FoV to match the raw data.
Non-squared FoV has larger view range than the squared FoV, thus is harder.
The input and novel views in fine-tuning are both 64 to support better view coverage.
The curriculum of the fine-tuning resolution is set as $128 \times 72$, $256 \times 144$, $512 \times 288$, and $960 \times 536$.
The training tokens for each stage is around $100$B.
High-resolution models (starting from $512 \times 288$) are trained with inner-chunk context parallelism (Sec.~\ref{sec:parallelism}). 

At each training stage, we always use AdamW with linear learning rate warmup and weight decay of $0.05$.
The peak learning rate of the pre-training is $4e-4$.
During fine-tuning, we use smaller learning rate (usually $1e-5$ to $5e-5$).

The training is completed with 64 A100 GPUs.
The pre-training takes $8$ days, and each fine-tuning stage is about $12$hours (thus 2 days in total).

\noindent\textbf{Detailed Result Numbers}
We here provided the detailed number for object-level  results on the GSO dataset (at resolution $256\times256$ in Table~\ref{tab:gso_256}, $512\times512$ in Table~\ref{tab:gso_512}) and DL3DV evaluations (at resolution $960\times536$ in Table~\ref{tab:dl3dv_960p}).


\begin{table}[h]
\centering
\caption{256-Res object-level novel view synthesis results on GSO. Both the input and output are with resolution $256\times256$. 
$\uparrow$: higher is better, $\downarrow$: lower is better.}
\resizebox{\textwidth}{!}{%
\begin{tabular}{lcccccccccc}
\toprule
\textbf{Input} & \textbf{\# Input Tokens} 
& \multicolumn{3}{c}{\textbf{LaCT}} 
& \multicolumn{3}{c}{\textbf{Full Attention}} 
& \multicolumn{3}{c}{\textbf{Perceiver Attention}} \\
\textbf{Views} & 
& \textbf{PSNR}~($\uparrow$) & \textbf{LPIPS}~($\downarrow$) & \textbf{SSIM}~($\uparrow$) 
& \textbf{PSNR}~($\uparrow$) & \textbf{LPIPS}~($\downarrow$) & \textbf{SSIM}~($\uparrow$) 
& \textbf{PSNR}~($\uparrow$) & \textbf{LPIPS}~($\downarrow$) & \textbf{SSIM}~($\uparrow$) \\
\midrule
4  & 4{,}096   & 32.4 & 0.030 & 0.962 & 32.6 & 0.029 & 0.964 & 30.3 & 0.039 & 0.950 \\
8  & 8{,}192   & 35.3 & 0.019 & 0.976 & 35.6 & 0.018 & 0.978 & 32.8 & 0.026 & 0.967 \\
12 & 12{,}288  & 36.3 & 0.017 & 0.980 & 36.6 & 0.015 & 0.982 & 33.6 & 0.023 & 0.971 \\
20 & 20{,}480  & 37.2 & 0.015 & 0.982 & 37.5 & 0.014 & 0.984 & 34.3 & 0.021 & 0.974 \\
32 & 32{,}768  & 37.5 & 0.014 & 0.982 & 37.9 & 0.013 & 0.985 & 34.2 & 0.021 & 0.974 \\
48 & 49{,}152  & 37.6 & 0.014 & 0.983 & 37.9 & 0.013 & 0.985 & 33.7 & 0.022 & 0.972 \\
\bottomrule
\end{tabular}
}
\label{tab:gso_256}
\end{table}

\begin{table}[h]
\centering
\caption{512-Res object-level novel view synthesis results on GSO. Both the input and output are with resolution $512\times512$ comparison across methods. $\uparrow$: higher is better, $\downarrow$: lower is better.}
\resizebox{0.85\textwidth}{!}{%
\begin{tabular}{lccccccc}
\toprule
\multirow{2}{*}{\textbf{Input Views}}
& \multirow{2}{*}{\textbf{\# Input Tokens}}
& \multicolumn{3}{c}{\textbf{LaCT}} 
& \multicolumn{3}{c}{\textbf{Full Attention}} \\
& 
& \textbf{PSNR}~($\uparrow$) & \textbf{LPIPS}~($\downarrow$) & \textbf{SSIM}~($\uparrow$) 
& \textbf{PSNR}~($\uparrow$) & \textbf{LPIPS}~($\downarrow$) & \textbf{SSIM}~($\uparrow$) \\
\midrule
4   & 16{,}384   & 33.4 & 0.029 & 0.969 & 33.6 & 0.027 & 0.971 \\
8   & 32{,}768   & 36.6 & 0.020 & 0.979 & 36.7 & 0.017 & 0.982 \\
12  & 49{,}152   & 37.7 & 0.017 & 0.983 & 37.9 & 0.015 & 0.985 \\
20  & 81{,}920   & 38.6 & 0.016 & 0.984 & 38.9 & 0.013 & 0.987 \\
32  & 131{,}072  & 39.0 & 0.015 & 0.985 & 39.3 & 0.013 & 0.988 \\
48  & 196{,}608  & 39.1 & 0.015 & 0.985 & 39.3 & 0.012 & 0.988 \\
\bottomrule
\end{tabular}
}
\label{tab:gso_512}
\end{table}

\begin{table}[h]
\centering
\caption{960P scene-level novel view synthesis results for LaCT on DL3DV. Both the input and output are with resolution $960\times536$ (width x height). $\uparrow$: higher is better, $\downarrow$: lower is better.}
\resizebox{0.55\textwidth}{!}{%
\begin{tabular}{lcccc}
\toprule
\textbf{Input Views} 
& \textbf{\# Input Tokens} 
& \textbf{PSNR}~($\uparrow$) 
& \textbf{LPIPS}~($\downarrow$) 
& \textbf{SSIM}~($\uparrow$) \\
\midrule
16  & 128{,}640  & 24.7 & 0.224 & 0.793 \\
32  & 257{,}280  & 26.9 & 0.185 & 0.837 \\
64  & 515{,}520  & 28.3 & 0.169 & 0.857 \\
128 & 1{,}031{,}520 & 28.9 & 0.166 & 0.861 \\
\bottomrule
\end{tabular}
}
\label{tab:dl3dv_960p}
\end{table}


\subsection{Language Modeling}\label{sec:appen_exp_language}

\noindent\textbf{Datasets \& Metrics.}
We train our models on the Long-Data-Collections dataset~\cite{long_data_collection}, containing approximately 68.8B tokens tokenized using Mixstra tokenizer (32,000 codebook size).
The dataset is a mix of 41.4\% General Data (e.g., RedPajama-Book, RedPajama-ArXiv, 1B tokens from RedPajama, and a Pile subsample) and 58.6\% Instruction Data (e.g., UL2 Oscar, NI, and P3). 
To evaluate long-context capabilities, we utilized the per-token loss metric from~\cite{linforgetting}. A consistently decreasing per-token loss across the input sequence indicates effective use of the entire context, while a plateau suggests an inability to leverage information beyond that point. Specifically, we evaluated next-token prediction loss on 2.5B tokens from the Book-3 dataset~\cite{gao2020pile} for our 760M-parameter model, and on 5B tokens for our 3B-parameter model. Additionally, we measured retrieval accuracy using the RULER benchmark~\cite{hsiehruler} across various sequence lengths to assess context memorization and information retrieval, evaluating up to the trained sequence length.

\noindent\textbf{Model details.}
We modified the original \methodname{} block by removing its window-attention layer. Instead, we incorporate a causal sliding window-attention(SWA) layer directly into the Large-Chunk TTT layer. The SWA layer shares the same Q, K, and V vectors
as the fast-weight network, except that a per-channel leranable scale and shift is applied  
to Q and K before they are fed to the SWA layer (as done in GAU~\cite{hua2022transformer}).  We sum up the output of the SWA layer and that of the TTT layer, where the output of the TTT layer is scaled by 
another per-head learnable scalar. 
We use an fast-weight lr initialization of $0.001$ by setting `const\_lr\_bias' in Algorithm~\ref{alg:full_lact_layer} to $\mathrm{softplus}(\text{const\_lr\_bias}) = 0.001$.
We illustrate this archtecture in Figure~\ref{fig:appen_lm_model}. 

To ensure a fair comparison with baselines in terms of trainable parameters.we adjusted the \methodname{} block's extra learnable initial fast weights $W = [W_1, W_2, W_3]$. o reduce parameters, we employed a low-rank version for $W_1, W3$ with a rank of 32. For instance, if $W_1 \in \mathcal{R}^{d\times d}$, its low-rank initial fast weight is $W_1 = L \cdot R + 0.5 * I_d$, where $L \in \mathcal{R}^{d\times32}, R \in \mathcal{R}^{32\times d}$, and $I_d$ is identity matrix. This reduces the extra trainable parameters for the fast weights in each block to $128 * \text{model-dim} + \frac{1}{\text{num-heads}} (\text{model-dim}^2) $. Additional minor parameters for learning rate projection, per-head scalers, an extra RMSNorm, and the SWA's learnable scale and shift are of order $O(\text{model-dim})$.  Standard blocks typically have $12 \cdot \text{model-dim}^2$ parameters. Our approach adds approximately $\frac{1}{\text{num-heads}} (\text{model-dim}^2)$ extra parameters, which is less than 3\% of total trainable weights with four heads, and below 4.5\% with two heads.  By default, \methodname{} use four heads in the experiments, unless noted otherwise, which means that the default state size per block is $\frac{3}{4} (\text{model-dim}^2)$.

\noindent\textbf{Baselines.}
We compare our approach with full attention, Gated Linear Attention (GLA)~\cite{yang2024gated}, DeltaNet~\cite{schlag2021linear,yangparallelizing}. To ensure fairness, we enhance both GLA and DeltaNet with the same sliding window attention.  As pointed out in previous work \cite{linforgetting, xiong2023effective, men2024base}, a large RoPE~\cite{su2023roformer} base is critical for transformers in long-context training, thus we adopt a large RoPE base of 1 million for training with 32K token contexts whenever softmax attention is used.
Tab.~\ref{tab:appen_LM_baseline} compares the mechanism and computing complexity of the baseline methods and our method. Training throughput (tokens per second per GPU, TPS) was using a 3B-parameter model on eight A100-40GB SXM4 GPUs with activation checkpointing and FSDP. At the 3 billion parameter scale, all models use 24 softmax attention heads.  The GLA baseline has eight linear attention heads with heads dimension as 384, resulting in a total state size of $384d$, with $d=3072$ representing the model dimension.  DeltaNet employs 24 linear attention heads, each with a dimension of 128, leading to a total state size of $128d$. Our approach uses four TTT heads with head dimension as $768$, and since each block has three fast weights, the total state size is $2304d$.

\begin{table*}[htbp]
\centering
\caption{Comparison of baseline methods in terms of state size, training throughput (measured in tokens per second, TPS), update rules, and memory read-out mechanisms. Training throughput is evaluated using a 3B-parameter model with 32K-sequence length on A100-40GB GPUs.}
\label{tab:appen_LM_baseline}
\resizebox{\textwidth}{!}{%
\begin{tabular}{llllll}
\toprule
 & \textbf{State size}  & \textbf{Train TPS} &  \textbf{Update Rule} & \textbf{Memory read-out} \\
\midrule
Transformer     & –       & 4.1K     & – & – \\
Transformer SWA & –       & 6.4K   & – & – \\
\midrule
\multicolumn{5}{l}{ \textit{Per-token recurrence} } \\
GLA SWA         & $384d$  & 5.0K   &
$\displaystyle \mathbf{S}_t \leftarrow  \mathbf{S}_{t-1}\mathrm{Diag}(\boldsymbol{\alpha}_t) + \mathbf{v}_t \mathbf{k}_t^\top$ 
& $\displaystyle \mathbf{o}_t = \mathbf{S}_t \mathbf{q}_t$ \\
DeltaNet SWA    & $128d$  & 5.1K   &
$\displaystyle \mathbf{S}_t \leftarrow  \mathbf{S}_{t-1}(\mathbf{I} - \beta_t \mathbf{k}_t \mathbf{k}_t^\top) + \beta_t \mathbf{v}_t \mathbf{k}_t^\top$ 
& $\displaystyle \mathbf{o}_t = \mathbf{S}_t \mathbf{q}_t$ \\
\midrule
\multicolumn{5}{l}{ \textit{Large-chunk recurrence} } \\
Ours GD       & $2304d$ & 5.0K    
& $ W \leftarrow \mathrm{L2norm}(W - \sum_i^b \eta_i \nabla_W \mathcal{L}_i)$
& $\displaystyle \mathbf{o}_t = f_W(\mathbf{q}_t)$  \\
Ours  Momentum       & $2304d$ & 4.9K    
& $M \leftarrow \beta M + \sum_i^b \eta_i \nabla_W \mathcal{L}_i;\; W \leftarrow  \mathrm{L2norm}(W - M)$
& $\displaystyle \mathbf{o}_t = f_W(\mathbf{q}_t)$  \\
Ours Muon       & $2304d$ & 4.3K  
& $ M \leftarrow  \beta M + \sum_i^b \eta_i \nabla_W \mathcal{L}_i;\;  W \leftarrow  \mathrm{L2norm}(W - \mathrm{Muon}(M))$
& $\displaystyle \mathbf{o}_t = f_W(\mathbf{q}_t)$ \\
\bottomrule
\end{tabular}%
}
\end{table*}

\noindent\textbf{Training Details.}
We trained models at two scales using a sequence length of 32,768 tokens:

\begin{itemize}
    \item 760M parameters: We use 24 stack blocks, with model dimension as $1536$. All models are trained for 40B tokens (40,960 steps) with a sliding window of 2048 tokens and a batch size of 1 million tokens. Each experiment ran on 32 A100-40GB SXM GPUs for approximately 20 hours.
    \item 3B parameters: We use 25 stack blocks, with model dimension as $3072$. All models are trained for 60B tokens (30,000 steps) with a sliding window of 4096 tokens and a batch size of 2 million tokens. Each experiment ran on 64 A100-40GB SXM GPUs for approximately 50-60 hours.
\end{itemize}
For both scales, we used a base learning rate of $1\times 10^{-3}$ 
 with a cosine decay scheduler and 1024 warmup steps. All models were randomly initialized with a standard deviation of 0.02.

\noindent\textbf{Results.} Detailed results on the RULER benchmark~\cite{hsiehruler} are presented in Tables~\ref{tab:ruler-single} and ~\ref{tab:ruler-multi}. We evaluated models on S-NIAH-1, S-NIAH-2, and S-NIAH-3 tasks, which represent varying difficulties of the single "needle in a haystack" retrieval. We also report performance on NIAH-MultiKey-1, NIAH-MultiQuery, and NIAH-MultiValue. Other RULER tasks are not reported as the full attention baseline also achieved trivial results beyond a 16K sequence length.

\begin{table}[ht]
\centering
\caption{RULER benchmark results for Single Needle in a Haystack (S-NIAH) tasks. * Our method with two heads (default is four).}
\label{tab:ruler-single}
\resizebox{\textwidth}{!}{%
\begin{tabular}{lcccc|cccc|cccc|cccc}
\toprule
& \multicolumn{4}{c}{S-NIAH-1} & \multicolumn{4}{c}{S-NIAH-2} & \multicolumn{4}{c}{S-NIAH-3} & \multicolumn{4}{c}{Average} \\
\cmidrule(lr){2-5}\cmidrule(lr){6-9}\cmidrule(lr){10-13}\cmidrule(lr){14-17}
Model & 4K & 8K & 16K & 32K & 4K & 8K & 16K & 32K & 4K & 8K & 16K & 32K & 4K & 8K & 16K & 32K \\
\midrule
\multicolumn{17}{l}{\textit{760M parameters}} \\
Transformer & 99.2 & 96.6 & 85.2 & 68.0 & 100 & 100 & 85.8 & 82.2 & 81.0 & 73.8 & 74.8 & 36.8 & 93.4 & 90.1 & 81.9 & 62.3 \\
DeltaNet + SWA & 84.0 & 85.2 & 87.8 & 86.8 & 62.8 & 29.4 & 14.2 & 7.8 & 53.8 & 21.8 & 11.2 & 5.8 & 66.9 & 45.5 & 37.7 & 33.5 \\
GLA + SWA & 51.8 & 26.2 & 14.4 & 8.6 & 55.8 & 26.4 & 15.8 & 7.8 & 58.0 & 23.8 & 16.2 & 5.0 & 55.2 & 25.5 & 15.5 & 7.1 \\
Ours & 94.8 & 53.2 & 26.0 & 14.8 & 74.0 & 28.0 & 14.2 & 7.8 & 42.8 & 26.6 & 14.4 & 6.8 & 70.5 & 35.9 & 18.2 & 9.8 \\
Ours Momentum & 95.6 & 84.8 & 83.4 & 84.8 & 91.4 & 73.4 & 22.8 & 7.8 & 82.6 & 34.8 & 16.6 & 6.6 & 89.9 & 64.3 & 40.9 & 33.1 \\
Ours Momentum* & 59.0 & 30.0 & 12.4 & 8.4 & 93.4 & 50.0 & 18.2 & 7.8 & 60.2 & 25.6 & 14.2 & 6.8 & 70.9 & 35.2 & 14.9 & 7.7 \\
Ours Muon & 98.0 & 95.0 & 92.2 & 92.4 & 86.6 & 60.2 & 17.0 & 7.8 & 49.2 & 26.2 & 10.9 & 5.2 & 77.9 & 60.5 & 40.0 & 35.1 \\
\midrule
\multicolumn{17}{l}{\textit{3B parameters}} \\
Transformer & 100 & 100 & 100 & 100 & 100 & 99.8 & 100 & 98.6 & 98.6 & 95.8 & 90.8 & 75.0 & 99.5 & 98.5 & 96.9 & 91.2 \\
GLA SWA & 100 & 52.8 & 26.0 & 13.2 & 100 & 51.8 & 29.6 & 14.4 & 98.0 & 54.4 & 27.6 & 12.4 & 99.3 & 53.0 & 27.7 & 13.3 \\
DeltaNet SWA & 100 & 89.6 & 76.2 & 54.8 & 100 & 76.4 & 42.2 & 17.0 & 90.6 & 57.6 & 27.4 & 13.4 & 96.9 & 74.5 & 48.6 & 28.4 \\
Ours Momentum & 99.4 & 97.0 & 98.6 & 93.4 & 100 & 75.6 & 39.6 & 15.0 & 91.8 & 63.0 & 27.8 & 13.4 & 97.1 & 78.5 & 55.3 & 40.6 \\
Ours Muon & 98.8 & 99.2 & 98.6 & 93.4 & 100 & 99.0 & 83.2 & 30.8 & 95.4 & 90.8 & 55.6 & 19.8 & 98.1 & 96.3 & 79.1 & 48.0 \\
\bottomrule
\end{tabular}%
}
\end{table}

\begin{table}[ht]
\centering
\caption{Performance on Multi-Key (MK-NIAH), Multi-Query (MQ-NIAH), and Multi-Value (MV-NIAH) Needle in a Haystack tasks from the RULER benchmark. * Our method with two heads (default is four).}
\label{tab:ruler-multi}
\resizebox{\textwidth}{!}{%
\begin{tabular}{lcccc|cccc|cccc|cccc}
\toprule
 & \multicolumn{4}{c}{MK-NIAH} 
 & \multicolumn{4}{c}{MQ-NIAH} 
 & \multicolumn{4}{c}{MV-NIAH} 
 & \multicolumn{4}{c}{Average} \\
\cmidrule(lr){2-5}\cmidrule(lr){6-9}\cmidrule(lr){10-13}\cmidrule(lr){14-17}
Model 
 & 4K & 8K & 16K & 32K 
 & 4K & 8K & 16K & 32K 
 & 4K & 8K & 16K & 32K 
 & 4K & 8K & 16K & 32K \\
\midrule
\multicolumn{17}{l}{\textit{760M parameters}} \\
Transformer
 & 63.8 & 72   & 71.4 & 54    
 & 33.4 & 28.9 & 24   & 23.1  
 & 27.95 & 24   & 20.5 & 27.35
 & 41.7 & 41.6 & 38.6 & 34.8 \\
DeltaNet+SWA 
 & 41.2 & 30 & 14.6 & 8.2    
 & 33   & 22.45 & 7.5 & 4.3    
 & 32.4 & 22.8 & 9.15 & 6.6
 & 35.5 & 25.1 & 10.4 & 6.4 \\
GLA + SWA 
 & 45.4 & 28.4 & 15.8 & 6.6
 & 26.1 & 17.75 & 10.2 & 5.85
 & 25.4 & 16.85 & 10.1 & 6.6
 & 32.3 & 21.0 & 12.0 & 6.3 \\
Ours 
 & 60.8 & 34.6 & 16.8 & 7
 & 35   & 23.65 & 14.1 & 7.45
 & 20.7 & 22.05 & 12.7 & 6.85
 & 38.8 & 26.8 & 14.5 & 7.1 \\
Ours Momentum 
 & 62 & 41 & 21.2 & 10.4 
 & 35.3 & 24.95 & 17.7 & 8.6 
 & 27.9 & 23.15 & 16.65 & 8.2
 & 41.7 & 29.7 & 18.5 & 9.1 \\
Ours Momentum* 
 & 59.8 & 37.8 & 19.2 & 8.8
 & 36.65 & 20.45 & 12.5 & 7.4
 & 24.45 & 16.95 & 11.6 & 6.8
 & 40.3 & 25.1 & 14.4 & 7.7 \\
Ours Muon 
 & 62.8 & 46.6 & 22 & 8.6 
 & 37.7 & 26.55 & 15.7 & 7.1
 & 28.35 & 23.15 & 13.6 & 6.85
 & 42.9 & 32.1 & 17.1 & 7.5 \\
\midrule
\multicolumn{17}{l}{\textit{3B parameters}} \\
Transformer 
 & 95 & 90.4 & 81.6 & 65.2
 & 86.45 & 81.55 & 71.70 & 40.85
 & 61.8 & 42.8 & 30.75 & 22.9
 & 81.1 & 71.6 & 61.4 & 43.0 \\
GLA 3B  
 & 78 & 45.8 & 28.6 & 14.4
 & 50.05 & 28.05 & 19 & 10.7
 & 29.4 & 21.4 & 16.75 & 9.9
 & 52.5 & 31.8 & 21.4 & 11.7 \\
DeltaNet SWA 
 & 75.8 & 57.4 & 34.2 & 17.8
 & 66.25 & 33.05 & 21.45 & 13.45
 & 43.7 & 23.2 & 18.85 & 13.2
 & 61.9 & 37.9 & 24.8 & 14.8 \\
Ours Momentum 
 & 96.2 & 59.6 & 35 & 17.2 
 & 87.05 & 40.25 & 25.6 & 13.2  
 & 88.08 & 30.65 & 21.9 & 12.3 
 & 90.4 & 43.5 & 27.5 & 14.2 \\
Ours Muon 
 & 75.2 & 69.2 & 46.2 & 25.2
 & 44.75 & 39.1 & 24.9 & 19
 & 26.55 & 29.1 & 25.05 & 19.3
 & 48.8 & 45.8 & 32.0 & 21.2 \\
\bottomrule
\end{tabular}%
}
\end{table}

\subsection{Autoregressive Video Diffusion}
\label{sec:appen_exp_video}
We fine-tune the pretrained Wan 2.1~\cite{wang2025wan} text-to-video diffusion model into an autoregressive video diffusion model, that generates videos by iteratively denoising successive chunks of video frames.


\noindent\textbf{Model details.}  The original Wan 2.1 is a bidirectional diffusion transformer operating on the latent space of a causal video VAE, which performs 8x spatial and 4x temporal downsampling. The diffusion transformer uses a $2\times2\times1$ patchification layer to convert VAE video latents to tokens.   Each block of the diffusion transformer comprises an MLP layer, a bidirectional self-attention layer for visual tokens, and a cross-attention layer for visual and text tokens.

Our primary modification is to the bidirectional self-attention. We first replace it with block-causal sliding window attention (SWA), using a window size of two chunks of video frames. We then integrate our \methodname{} into the same layer. We initialize learnable fast weights for \methodname{}. Consistent with our language modeling experiments, SWA and our test-time training mechanism are combined within each layer: Q and K vectors are rescaled and shifted before input to the test-time training operation. The outputs of SWA and the test-time training layer are summed, with a per-head learnable scalar (from a zero-initialized linear projection) applied to the latter. We do not use Muon in the fast-weight update, as it showed no significant difference in validation loss empirically. 
We use an fast-weight lr initialization of $0.001$ by setting `const\_lr\_bias' in Algorithm~\ref{alg:full_lact_layer} to $\mathrm{softplus}(\text{const\_lr\_bias}) = 0.001$.
This allows small update to the fast weight in the beginning of the fine-tuning.  To maintain minimal changes to the original Wan architecture, \methodname{} layers utilize the original RoPE from the Wan model, and we remove the SiLU activation function previously applied to queries and values.
 

\noindent\textbf{Datasets.} We fine-tune the model using an internal, filtered proprietary collection of videos, each accompanied by a short text prompt generated by a visual language model\cite{chen2024internvl}.

\noindent\textbf{Training details.}
Following~\cite{esser2403scaling, wang2025wan}, we use time-step shifting (scale factor 3.0) and logit-normal denoising loss weighting (mean=0.5, std=1.0). We also apply an exponential moving average with a decay rate of 0.995 to the model weights.  Each 5-second video (16 FPS, 480×832 resolution) is encoded by the Wan VAE into a [21,60,104] latent representation. Denoising is performed autoregressively in chunks of three latent frames (4680 visual tokens each). We employ teacher-forcing with an interleaved noisy-clean chunk sequence (see Section~\ref{sec:method-arvideo}). 

\begin{itemize}
    \item \textbf{1.3B Parameter Model:} For initial training on 5-second videos, noisy chunks are repeated twice. This results in sequences of 60 latent frames (14 noisy, 6 clean chunks), totaling 93,600 tokens. We finetune the model with a batch size of 64 for 5000 iterations. The base learning rate is set to $2\times10^{-5}$ with a linear warm-up of 1000 iterations and linear decay.  Subsequently, the model is fine-tuned on 10-second video clips for 1,000 iterations. These clips correspond to 42 latent frames for the clean video portion, forming an interleaved sequence of 81 latent frames (approximately 126K tokens including noisy chunks).   Training for 5-second videos takes $\sim$20 seconds per iteration on 64 A100 80GB SXM GPUs (or $\sim$10 seconds on 64 H100 80GB SXM GPUs).
    \item  \textbf{14B Parameter Model:}. To manage GPU memory usage, noisy chunks are not repeated in this setting. We train the model on five-second videos with a batch size of 64 for 5000 iterations with a base learning rate of $5\times10^{-6}$, and use a sequence parallel size of 2 GPUs. This phase takes $\sim$80 seconds per iteration on 64 A100 GPUs. The model is then fine-tuned on 8.8-second video clips (36 latent frames for the clean portion) for an additional 600 iterations, using sequence parallelism (4 GPUs). This fine-tuning takes $\sim$80 seconds per iteration on 64 H100 GPUs.
\end{itemize}

\noindent\textbf{Baselines.}
We compare our method with three baselines: sliding window attention, Mamba2~\cite{dao2024transformers} with sliding window attention, and full block-wise causal attention, where the window attention in the baselines is implemented the same as in our model. 
For the Mamba2 layer, we follow \cite{wang2024mambainllama} to apply the original projected $k$, $q$, and $v$ as $B$, $C$, and $x$, respectively. The Mamba2's state is updated token-by-token, we revert the state after processing a noise chunk of frames to ensure only clean chunk state updates propagate.
The full block-wise causal attention baseline is implemented with FlexAttention~\cite{he2024flexattention}.

\noindent\textbf{Evaluation.} We validate all models on a collection of 2,000 videos after 5,000 training iterations by computing the denoising loss at five timesteps (550, 650, 750, 850, 950). 
The denoising losses are measured with respect to each video frame chunk and plotted in Figure~\ref{fig:exp-video-baseline}. Figure~\ref{fig:exp-video-baseline}(a) compares validation loss (up to 5s videos) of \methodname{} against SWA, Mamba2 with SWA, and full block-wise causal attention. Our \methodname{} is comparable to full attention and outperforms other baselines. Figure~\ref{fig:exp-video-baseline}(b) shows comparisons with the SWA baseline using different window sizes for both our method and the baseline (up to 5s videos). The default window covers six latent frames (two chunks). An additional experiment used a four-frame window. Results indicate that increasing window size from four to six frames improves validation loss, but this improvement is smaller than that achieved by incorporating \methodname{}. 
Figure~\ref{fig:exp-video-baseline}(c) presents validation loss (up to 10s videos) after fine-tuning \methodname{} and the SWA baseline on 10-second videos for 1,000 iterations.

Generated video samples from our model are provided in an appended folder. Each video chunk is sampled following the original Wan method, using a UniPC~\cite{zhao2023unipc} sampler with 50 steps, classifier-free guidance of 5.0, and a timestep shift of 3.0. 

\subsection{Experiment details in Figure 1}
\label{sec:appen_figure_1_details}

Fig.~\ref{fig:compare_with_ttt}(c) shows results for training a 760M-parameter \methodname{} language model. We employ a SwiGLU MLP fast weight with the Muon test-time optimizer. To scale the fast weight size, we fix the intermediate dimension of the fast-weight MLP to match the head dimension, then increase the head dimension from 128 to 1536 while proportionally decreasing the number of heads to maintain a constant total model dimension. Validation loss is computed on the last 2,048 tokens of each 32,768-token sequence, averaged over 76K sequences from the Book3 dataset.

Fig.~\ref{fig:compare_with_ttt}(d) uses the object-level novel view synthesis experiment. All models consist of 14 stacked blocks with a fixed model dimension of 768 and were trained for 167 billion tokens. Training time (wall-clock) is measured on an A100-40GB SXM GPU.

\section{Details for Mamba Baselines}

Mamba is an efficient model architecture, it is logically similar to a linear TTT taking per-token linear update rule of the fast weight (i.e., state in Mamba's context).
Thus it serves as a baseline to understand the gap between the chunk-wise update and per-wise update in Fig.~\ref{fig:exp-linear-token-recurrence} and Fig.~\ref{fig:exp-video-baseline}.
In this section, we detailed the experimental setup.

We take the official Mamba-2 implementation\footnote{\hyperlink{https://github.com/state-spaces/mamba}{https://github.com/state-spaces/mamba}} in all our experiment. The original Mamba-2 has multiple components, and we largely simplify its implementation to keep a measurable architecture while still maintaining the performance.
In detail, our Mamba-2's formulation in experiment is:
\begin{align}
    X, B, C, \delta & = \mathrm{Linear}(u) \nonumber \\
    \delta & = \mathrm{softplus}(\delta + \delta_\mathit{init}) \nonumber \\
    H_t &= \exp(-\delta_t) H_{t-1} + \delta_t B_t^T X_t \nonumber \\
    y_t &= C_t H_t 
\end{align}
where $X$, $B$, $C$ is of shape $(L, d)$, and $\delta$ is of shape $(L, 1)$. $H_t$ is a matrix state of shape $(d, d)$.

Transferring the above formula to a standard linear-attention / TTT / DeltaNet notations, it is equivalent to:
\begin{align}
    V, K, Q, \mathit{lr} & = \mathrm{Linear}(\mathit{input}) \nonumber \\
    \mathit{lr} & = \mathrm{softplus}(\mathit{lr} + \mathit{lr}_\mathit{init}) \nonumber \\
    W_t &= \exp(-\mathit{lr}_t) W_{t-1} + \mathit{lr}_t K_t^T V_t   \nonumber \\
    O_t &= Q_t W_t 
    \label{eqn:mamba2}
\end{align}
We will denote the above equations as $O = \mathrm{Mamba}(\mathit{input})$.

We use the multi-head design as in Transformer's multi-head attention.
Multiple independent Mamba-2 layer are run in parallel and their outputs are concatenated.
Suppose the number of heads is $\mathit{nh}$, the formula is:
\begin{align}
    O^k & = \mathrm{Mamba^k} (\mathit{input}) \nonumber \\
    O &= [O^1, \ldots, O^\mathit{nh}]
\end{align}
where each $\mathrm{Mamba^k}$ is a Mamba with its own parameters.
In Mamba-2's terminology, this design is equivalent to setting the number of `groups' to be the same as the number of heads.

For the novel view synthesis task, we take a bidirectional Mamba over the input image tokens.
In detail, we take two independent multi-head Mamba with one reading from left to right and the other reading from right to left.
The bidirectional model builds a better connection among input tokens and also doubles the state size.
We use a similar `apply' operation as in \methodname{} that only updates the state for input tokens, and the state is static for the target tokens.
We also tested with `update' for the target image tokens, but it empirically leads to worse results.
We use a head dimension of $192$ and $8$ heads.
The overall state size, $8~\text{(num heads)} \times 192^{2} ~\text{(head dim)} \times 2~\text{(bidir)}$, matches \methodname{} with a standard large-chunk large-weight linear attention of dimension $768$ ($768~\text{input dim} \times 768~\text{intermediate dim}$ in Fig.~\ref{fig:exp-linear-token-recurrence}.
We take $\mathit{lr}_\mathit{init}{=}-4.6$, which corresponds to a $0.01$ initialized learning rate (i.e., $\mathrm{softplus}(-4.6) = 0.01$).

For the autoregressive video diffusion task, we apply a unidirectional Mamba over the flattened video tokens. As mentioned in Sec~\ref{sec:exp_video}, we follow~\cite{wang2024mambainllama} to inherit the Wan's self-attention projected $k$, $q$, and $v$ as $B$, $C$, and $x$ in the Mamba layer, respectively. Unlike in the NVS task, each token will `update' the state, which will be `applied' to the current output and future tokens. Our Mamba uses 12 heads, each of dimension 128, matching the original multi-head self-attention in Wan. The overall state size is $12~\text{(num heads)} \times 128^{2} ~\text{(head dim)}$.8*19
We take $\mathit{lr}_\mathit{init}{=}-4.6$, which corresponds to a $0.01$ initialized learning rate.

\section{Details of LaCT Context Parallelism Implementation}

Context Parallelism(CP) partitions the input sequence along its sequence length dimension and distributed the shards across multiple devices for parallel computing. The feed-forward layer and window attentions are local operations thus support CP naively.  Our large-chunk Test-Time Training (TTT) approach facilitates CP by sharding tokens within each large chunk.

Within our large-chunk TTT mechanism, the per-token \textit{apply} operation naively supports CP due to its independent nature. The \textit{update} allows CP by shading tokens within a chunk over multiple devices. This CP can be easily implemented by adding a few lines of distributed all-reduce-sum after computing the local fast weight gradients on each device, logically the same as the Distributed Data Parallellism.  
Note that the distributed all-reduce-sum is a differentiable operator and its backward is all-reduce-sum over the gradient, thus the network can be trained end-to-end.
Algorithm~\ref{alg:lact_chunk_cp} presents the pseudocode detailing this intra-chunk context parallelism specifically for the large-chunk TTT \textit{update} operation. We employed this parallelism in our view synthesis experiments, handling maximum chunk sizes exceeding half a million tokens and maximum sequence lengths over one million tokens during training.

\begin{algorithm}[t]

\caption{Large Chunk Test-Time Training Layer with Context Parallel Sharded inside chunk Pseudocode}\label{alg:lact_chunk_cp}
\newcommand{\hlbox}[1]{%
  \fboxsep=1.2pt\hspace*{-\fboxsep}\colorbox{blue!10}{\detokenize{#1}}%
}
\lstset{style=mocov3}
\begin{lstlisting}[
    language=python,
    escapechar=@,
    label=code:lact_chunk_cp]
    def update(fast_weight, k, v, lr, cp_group, use_muon=True):
        """
        Fast-weight update for a SwiGLU MLP using a context-parallel chunk. 

        Args:
            fast_weight : tuple(w1, w2, w3) with shapes: w1, w3: [b, d, dh]; w2: [b, dh, d]
            k, v        : key / value tensor of shape [b, l, d]
            lr:         : per-token learaning rates of shape [b, l, 3] -> (lr1, lr2, lr3)
            cp_group    : process group metadata for context parallelism
            use_muon    : weather to apply Muon to orthogonalize the update

        Note:
            The input tensors k, v, lr are assumed to be already partitioned (sharded) along the sequence 
            dimension over multiple devices. l represents the local sharded sequence length on each device.
            The total effective chunk size processed is  l * cp_group.size.
        """
        
        # Forward with k:
        gate_before_act = matmul(k, w1) # [b, l, dh] = [b, l, d] x [b, d, dh]
        hidden_before_gate = matmul(k, w3) # [b, l, dh] = [b, l, d] x [b, d, dh]
        hidden = silu(gate_before_act) * hidden_before_gate
        
        # Backward:
        dhidden = matmul(v, w2.transpose(-1, -2)) # [b, l, dh] = [b, l, d] x [b, d, dh]
        dhidden_before_gate = dhidden * silu(gate_before_act)
        dgate = dhidden * hidden_before_gate
        dgate_before_act = silu_backprop(dgate, gate_before_act)

        # Compute gradients: 
        w2.grad = -matmul(hidden.transpose(-1, -2), v * lr2) # [b, dh, d] = [b, dh, l] x [b, l, d]
        # [b, d, dh] = [b, d, l] x [b, l, dh]
        w1.grad = -matmul((k * lr1).transpose(-1, -2), dgate_before_act)
        w3.grad = -matmul((k * lr3).transpose(-1, -2), dhidden_before_gate)

        # [Standard forward pass and local backward gradient computations are performed above,
        #  resulting in local w.grad for each device.]

        ####################################################################################
        # BEGIN CONTEXT PARALLELISM SPECIFIC MODIFICATION: Global Gradient Aggregation      
        # The following AllReduce operation is the key step introduced for context             
        # parallelism. Operations before this point compute local gradients; operations   
        # after this point use the globally aggregated gradients.            
        ####################################################################################
        for w in fast_weight:
            w.grad = distributed_all_reduce(w.grad, cp_group, op="SUM")
        ####################################################################################
        # END CONTEXT PARALLELISM SPECIFIC MODIFICATION.                 
        # Subsequent operations (Muon, weight updates) now use the globally summed w.grad.
        # The formulas for these subsequent operations remain the same as in a 
        # non-parallel version, but they act upon these aggregated gradients.   
        ####################################################################################

        # Weight update
        if use_muon:
            for w in fast_weight:
                w.grad = zeropower_via_newtonschulz5(w.grad)
        for w in fast_weight:
            w = (w - w.grad) / (w - w.grad).norm(dim=1) * w.norm(dim=1)

        return fast_weight
\end{lstlisting}
\end{algorithm}

\section{Details of LaCT Tensor Parallelism Implementation}

Beyond Context Parallelism, our large-chunk Test-Time Training (TTT) mechanism also supports Tensor Parallelism (TP). This is primarily achieved by sharding the TTT heads across multiple devices, a strategy similar to that employed in methods like DeepSpeed Ulysses~\cite{jacobs2023deepspeed}. 

Specifically, while static feed-forward layers in the model might process inputs sharded along the sequence dimension (Context Parallelism), for the TTT operations within our LaCT layer, the data undergoes a gather-then-scatter transformation. Input tensors (Q, K, V, and learning rates for TTT) that are initially sharded by sequence length are first gathered along the sequence dimension to reconstruct the full sequence context on each device within the tensor-parallel group. Then, these full-sequence tensors are scattered along the head dimension. As a result, each device processes the complete sequence but operates on only its assigned subset of TTT heads during the TTT \textit{update} and \textit{apply} iterations. The reverse transformation (gather heads, scatter sequence) is applied to the output of TTT operation. Algorithm~\ref{alg:lact_tp} provides pseudocode detailing this tensor parallelism implementation, omitting minor details like padding. While this gather-then-scatter method effectively enables head-sharded tensor parallelism, more sophisticated communication strategies~\cite{fang2024usp, jacobs2023deepspeed} could potentially be employed to further optimize communication overhead. 

We utilized this tensor parallelism strategy in our autoregressive video generation experiments, sharding, for example, four TTT heads across four local GPUs. This enabled us to train 14-billion-parameter diffusion models with sequence lengths exceeding 100K tokens.

\begin{algorithm}[h]
\caption{Large Chunk Test-Time Training Layer with Tensor Parallelism by sharding heads Pseudocode}\label{alg:lact_tp}
\newcommand{\hlbox}[1]{%
  \fboxsep=1.2pt\hspace*{-\fboxsep}\colorbox{blue!10}{\detokenize{#1}}%
}
\lstset{style=mocov3}
\begin{lstlisting}[
    language=python,
    escapechar=@,
    label=code:lact_tp]
    def gather_scatter(x, gather_dim, scatter_dim, process_group=cp_group):
        """
        Gathers tensor x along gather_dim across process_group,
        then scatters the result along scatter_dim to each device locally.
        Example: Transform [B, N_full, L_local, D] with gather_dim=2, scatter_dim=1
                 to [B, N_local, L_full, D] on each device.
        """
        x = all_gather(x, gather_dim, process_group)

        # Calculate slicing indices for the scatter operation
        local_rank, group_size = process_group.rank, process_group.size
        scatter_stride = x.size(scatter_dim) // group_size
        start_idx = local_rank * scatter_stride
        end_idx = (local_rank + 1) * scatter_stride

        # Slice the tensor to get the local shard for the current device
        x = slice_tensor(x, scatter_dim, start_idx, end_idx)
        return x

        
    ############### MultiHead LaCT Layer with Tensor Parallelism (sharding TTT heads) ###############
    # Input: 
    # x: input sequence sharded by sequeunce length (CP). Shape [b, l, d], b is the batch dim, l is local sequence length, d is model dimension. 
    # fast_weight: tuple of sharded initial fast weights, sharede among heads.  (w1, w2, w3); w1, w3 of shape [nh, d, dh], w2 of shape [nh, dh, d].  nh: number of local heads. 

    qkv = silu(LinearQKV(x)) # [b, l, d * 3]
    qkv = rearrange(qkv, `b l (nh hd) -> b nh l hd`, nh=num_heads).split(3, dim=-1)
    q, k = q / q.norm(-1), k / k.norm(-1)
    lr = softplus(LinearLR(x) + const_lr_bias) # [b, l, 3 * num_heads]
    lr = rearrange(lr, `b l (nh 3) -> b nh l 3`, nh=num_heads)

    ####################################################################################
    # BEGIN TENSOR PARALLELISM SPECIFIC TRANSFORMATION 
    # Gather along Sequence Length (dim 2), then Scatter along Head Dimension (dim 1).
    # Transforms [b, nh_full, l_local, X] -> [b, nh_local, l_full, X]
    # Each device now has the full sequence for a subset of heads.
    q, k, v, lr = map(lambda x: gather_scatter(x, gather_dim=2, scatter_dim=1), (q, k, v, lr))
    # END TENSOR PARALLELISM SPECIFIC TRANSFORMATION
    ####################################################################################

    # [b, nh_local, l_full, X]
    o_local_heads = ... # Placeholder for actual TTT computation on sharded heads
    o_local_heads = RMSNorm(o_local_heads) # per-head norm

    ####################################################################################
    # BEGIN TENSOR PARALLELISM SPECIFIC REVERSE TRANSFORMATION 
    # Gather along Head Dimension (dim 1), then Scatter along Sequence Dimension (dim 2).
    # Transforms [b, nh_local, l_full, X] -> [b, nh_full, l_local, X]
    # This reconstructs the full head dimension but shards sequence back.
    o = gather_scatter(o_local_heads, gather_dim=1, scatter_dim=2)
    # END TENSOR PARallelism SPECIFIC REVERSE TRANSFORMATION
    ####################################################################################
    o = rearrange(o, `b nh l hd -> b l (nh hd)`, nh=num_heads)
    o = LinearOutput(o)

    return o
\end{lstlisting}
\end{algorithm}

\end{document}